\pdfoutput=1

\documentclass[11pt]{article}

\usepackage[]{coling}

\usepackage{times}
\usepackage{latexsym}

\usepackage[T1]{fontenc}

\usepackage[utf8]{inputenc}

\usepackage{microtype}

\usepackage{inconsolata}

\usepackage{graphicx}
\usepackage{xcolor,colortbl} 
\usepackage{microtype}
\usepackage{inconsolata}
\usepackage{tabularx}
\usepackage{hyperref}
\usepackage{amsmath}
\usepackage{mathtools}
\usepackage{todonotes}
\usepackage{hhline}
\usepackage{soul}
\usepackage{nicematrix}
\usepackage{bbm}
\DeclareRobustCommand{\redst}[1] {{\setstcolor{red}\st{#1}}}

\DeclareRobustCommand{\bluest}[1]{{\setstcolor{blue}\st{#1}}}

\DeclareRobustCommand{\hlgreen}[1]{{\sethlcolor{lightgreen}\hl{#1}}}

\usepackage{dashrule}
\usepackage{ulem}
\usepackage{cancel}

\usepackage{nicematrix}
\usepackage{times}
\usepackage{latexsym}
\usepackage{multirow}
\usepackage{sidecap}
\usepackage{amsmath}
\usepackage{amssymb}
\usepackage{amsfonts}
\usepackage{verbatim}
\usepackage{url}
\usepackage{pifont}
\usepackage{eqparbox}
\usepackage{comment}
\usepackage{comment}
\usepackage{graphicx}
\usepackage{booktabs}
\usepackage{enumitem}
\usepackage{caption}
\usepackage{subcaption}
\usepackage{bm}
\usepackage{makecell}

\definecolor{DarkRed}{RGB}{130,25,0}
\definecolor{DarkGreen}{RGB}{30,130,30}
\definecolor{DarkBlue}{RGB}{0,0,250}
\definecolor{DarkYellow}{RGB}{255,128,0}
\definecolor{lightyellow}{RGB}{250,253,191}
\definecolor{DarkGray}{RGB}{96, 96, 96}
\definecolor{lightgreen}{RGB}{229,255,204}

\newcommand{\xmark}{\textcolor{red}{\ding{55}}} 
\newcommand{\cmark}{\textcolor{DarkGreen}{\ding{51}}}
\newcommand{\addmark}{\textcolor{DarkYellow}{\ding{59}}}
\newcommand{\editmark}{\textcolor{blue}{\ding{34}}}
\newcommand{\sep}{-0.1cm} 

\newcommand{\interalia}[1]{\citep[\textit{i.a.}]{#1}} 

\definecolor{pale_green}{rgb}{0.55,0.75,0.60}
\definecolor{pale_red}{rgb}{0.90,0.61,0.58}
\definecolor{pale_yellow}{rgb}{0.95,0.92,0.72}

\title{Exploring the Reliability of Large Language Models as Customized Evaluators for Diverse NLP Tasks}

\author{ 
Qintong Li$^{\heartsuit}\footnotemark[1] $\hspace{0.5mm} \quad
Leyang Cui$^{\clubsuit}$ \quad 
Lingpeng Kong$^{\heartsuit}$ \quad
 Wei Bi$^{\clubsuit}$\hspace{0.5mm} \\
$^\heartsuit$ The University of Hong Kong \quad
$^\clubsuit$ Tencent AI lab 
 \\
\texttt{\{qtli,lpk\}}@cs.hku.hk\\
\quad\texttt{\{leyangcui,victoriabi\}@tencent.com} \\
}

\begin{document}
\maketitle
\renewcommand{\thefootnote}{\fnsymbol{footnote}}
\footnotetext[1]{Work was done during the internship at Tencent AI lab.}
\renewcommand*{\thefootnote}{\arabic{footnote}}

\begin{abstract}
Previous work adopts large language models (LLMs) as evaluators to evaluate natural language process (NLP) tasks. 
However, certain shortcomings, e.g., fairness, scope, and accuracy, persist for current LLM evaluators.
To analyze whether LLMs can serve as reliable alternatives to humans, we examine the fine-grained alignment between LLM evaluators and human annotators, particularly in understanding the target evaluation tasks and conducting evaluations that meet diverse criteria.
This paper explores both conventional tasks (e.g., story generation) and alignment tasks (e.g., math reasoning), each with different evaluation criteria.
Our analysis shows that 1) LLM evaluators can generate unnecessary criteria or omit crucial criteria, resulting in a slight deviation from the experts.
2) LLM evaluators excel in general criteria, such as fluency, but face challenges with complex criteria, such as numerical reasoning.
We also find that LLM-pre-drafting before human evaluation can help reduce the impact of human subjectivity and minimize annotation outliers in pure human evaluation, leading to more objective evaluation.
All resources are available at \url{https://github.com/qtli/CoEval}.
\end{abstract}

\section{Introduction}

The success of Large Language Models (LLMs) in executing real-world tasks according to instructions~\citep{AutoGPT,liu2023agentbench} has spurred increased interest in using LLMs as evaluators, with task examples treated as specific instructions during evaluation~\interalia{liu2023gpteval,zhang2023gpt,zheng2023judging}.
However, current LLM evaluators still have certain shortcomings. For example, an LLM evaluator for math reasoning tasks may not penalize deceptive solutions that contain incorrect steps~\cite{toh2023veritymath}. 
This issue is particularly severe for those intricate tasks where meticulous verification or logical reasoning are the main evaluation criteria~\citep{ling2023deductive,zeng2023evaluating}.
There is an urgent need for a systematic investigation on the reliability of LLMs as trustworthy and universal evaluators capable of replacing humans in various NLP tasks.

Investigating \textit{whether LLMs are capable of generating various yet adequate evaluation criteria across different tasks} is a necessary first step, so that we can understand the extent of agreement between LLM evaluators and humans in interpreting the instruction of ``evaluating a task''.
Furthermore, with the evolving nature of NLP tasks, it is important for LLM evaluators to flexibly meet new requirements and diverse criteria. 
This raises the pertinent question of \textit{whether the evaluation results of LLMs can be trusted and aligned with specific criteria}. If the answer is negative, is there a way to enhance the LLM evaluators?

We aim to provide separate and clear responses to both questions.
To gauge the task comprehension ability of an LLM evaluator, we prompt an LLM to offer a variety of evaluation criteria for the assigned tasks, followed by an examination of how well the LLM's criteria align with human expertise.
We consider three benchmarks, including question answering, story generation, and math word problem-solving, as well as 252 instruction-following tasks, across 692 distinct criteria that experts manually curate.
We find that LLMs can generate mostly consistent, valid and sufficient task-specific evaluation criteria. 
LLMs occasionally overlook critical criteria, such as ``conciseness'' for writing a brief report, which experts would prioritize (Section~\ref{sec:criteria-align-with-human}). This oversight could introduce biases in the subsequent sample evaluation.

To measure the evaluation quality of LLMs, we discuss several methods for instructing an LLM evaluator to score samples of a particular task, with or without evaluation criteria.
Meta-evaluation shows that when LLMs thoroughly consider criteria before summarizing an overall score, they can achieve a higher correlation with human assessments than when they directly evaluate.

Currently, LLMs still have a long way to go before they can replace humans.
For example, in evaluating the ``analogy usage'' on the long-form QA task, LLM tends to hallucinate and generate more positive scores; and in evaluating the ``logical reasonability'' on the math reasoning task,  the LLM evaluator easily fails to detect simple logical errors (Section~\ref{exp:llm-evaluator}).
We further explore the potential for an LLM to serve as a cost-effective auxiliary to human annotators by allowing humans to edit the results generated by the LLM evaluator (Section~\ref{exp:llm_human_eval}).
Compared with human-only evaluation, the inter-annotation agreement of Krippendorff's $\alpha$ notably improves from 0.64 to 0.71.
However, directly replacing pure human evaluation with this method may have some potential risks. 
LLM evaluations may inhibit certain aspects of human subjectivity while mitigating certain annotation outliers.

Based on the efforts and outcomes of this study, we encourage further research on LLMs as evaluators. This includes exploring their usage in challenging new tasks, designing proper prompts for evaluation, and conducting rigorous tests to assess their trustworthiness as evaluators for the task.

\section{Related Work}

Traditional automatic metrics are well established for judging specific NLP tasks, such as BLEU~\citep{papineni2002bleu} for machine translation, ROUGE~\citep{lin2004rouge} and METEOR~\citep{banerjee2005meteor} for text summarization, and CIDEr~\citep{vedantam2015cider} for image captioning.
To improve the correlation with human judgments, several approaches integrate contextual word embeddings, including MoverScore~\citep{zhao2019moverscore}, Sentence Mover's Similarity~\citep{clark2019sentence}, BERTscore~\citep{zhangbertscore}, and Bartscore~\citep{yuan2021bartscore}. 
Other related works propose task-specific metrics to align specific human assessments, e.g., consistency~\citep{durmus2020feqa,honovich2021q2,fabbri2022qafacteval}, coherence~\citep{durmus2020feqa,ye2021towards}, and grammar~\citep{pratapa2021evaluating}.
However, no universal metric exists that can accommodate all generation tasks and capture all desirable properties of language~\citep{reiter2009investigation,garbacea2020neural}. Human evaluation is prevalent in generation tasks~\citep{mathur2020tangled,belz2020disentangling,liu2023evaluating}.

As research in LLMs continues to accelerate, LLM-based
evaluation has emerged as a scalable and cost-effective alternative to human evaluations~\citep{jain2023multi,alpaca,vicuna2023,wu2023style}.
~\citet{fu2023gptscore} uses LLM's predicted text probability as the automated score.
Along a more black-box line, the community~\citep{wang2023chatgpt,chiang2023can,zeng2023evaluating,zhang2023wider,zhang2023gpt} has turned to induce LLM to directly generate evaluation scores for diverse tasks, such as summarization~\citep{liu2023gpteval} and dialogue~\citep{zheng2023judging}, with superior human correlation compared to conventional metrics. 
Prior studies primarily focus on devising efficient prompting strategies to elicit high correlations between LLMs and human annotators, reaching conclusions regarding the efficacy of LLMs as evaluators in a straightforward manner. 
Different from previous work, we not target any particular new task. 
Rather, our primary focus lies in how well LLMs align with human experts in understanding evaluation tasks and evaluating samples based on specific criteria.

\section{Evaluation Setup}

\paragraph{Tasks}
We consider three benchmarks and 252 instruction-following tasks that exhibit distinct characteristics and necessitate distinct criteria for evaluating their output. 
Investigating the behaviors of the LLM evaluator on these tasks enables us to derive broad observations.
Specifically, these tasks include:
(i) a long-formed QA task ELI5~\citep{eli5_lfqa} where the main focus is on the \textit{factuality} and \textit{comprehensibility} of the generated answers; 
(ii) a story generation task ROCStories~\citep{mostafazadeh2016corpus} that primarily emphasizes on the \textit{coherence} and \textit{relevance} and other quality aspects of the generated story; 
(iii) a math word problem task GSM8K~\citep{cobbe2021training} that people are often concerned with its reasoning ability, such as \textit{logic} and \textit{correctness};
(iv) an instruction-following dataset Self-Instruct~\citep{wang2022self}, which involves various daily scenarios (e.g., email writing and film review), and thus we may require distinct evaluation perspectives for different instructions within this dataset.

\paragraph{Evaluation Configurations}

This paper considers \texttt{gpt-3.5-turbo} the representative LLM evaluator due to its efficacy and economic benefits.
We also examine a more powerful model, i.e., \texttt{gpt-4}, when conducting evaluation, and the results are shown in Table \ref{tab:gpt4_result}.
For the benchmark tasks (ELI5, ROCStories, and GSM8K), the LLM evaluator and human annotators are instructed to assess the quality of outputs from three generation models\footnote{\url{{https://platform.openai.com/docs/models}}} (\texttt{gpt-3.5-turbo}, \texttt{text-davinci-002}, and \texttt{text-curie-001}), as well as human-written ground truth.
Each task includes 200 evaluation samples derived from 50 randomly selected inputs.
For Self-Instruct, we use the model outputs and human evaluations provided by its authors\footnote{\url{https://github.com/yizhongw/self-instruct/tree/main/human_eval}} because different tasks require different expertise, and it is extremely challenging to find qualified human annotators.

\section{LLM-Generated Criteria}
\label{sec:criteria_gen}

We first investigate the divergence between LLM evaluators and human experts in explaining the evaluated tasks and examine if an LLM evaluator can generate various yet adequate evaluation criteria for various tasks.

\subsection{Prompt for Evaluation Criteria}
\label{sec:prompt_for_criteria}
Given the substantial ability of \texttt{gpt-3.5-turbo} to adhere to directions, we utilize the prompt below to request it to generate evaluation criteria based on the task description and a task example.
The example includes the input $\boldsymbol{x}$ and an output $\boldsymbol{y}$.

\begin{quote}
\small
\textit{Now, we have a task \texttt{[$task$ $desc.$]}.}

\textit{Here is a demonstration example of the task:}

\texttt{Input: [$\boldsymbol{x}$]}
\texttt{Output: [$\boldsymbol{y}$]}

\textit{Please make sure you read and understand how to do this task.}

\textit{But your real task is to tell me how to evaluate this task. The evaluation criteria should include general criteria used in natural language tasks, as well as task-specific criteria about this evaluated task. Please provide a clear and comprehensive list of your evaluation criteria.}

\textit{Evaluation Criteria:}
\end{quote}

\noindent
For benchmark datasets, \texttt{[$task$ $desc.$]} is written by human experts. For the Self-Instruct task, \texttt{[$task$ $desc.$]} represents each instruction, such as ``Change the response to have a more empathetic tone in the chat.''.
The demonstrated input \texttt{[$\boldsymbol{x}$]} and output \texttt{[$\boldsymbol{y}$]} are randomly selected from the task dataset.
As \texttt{gpt-3.5-turbo} possesses strong instruction-following abilities, it can easily understand the instruction to output a set of criteria for a specific task based on the properties with 100\% completion. Therefore, our study mainly focuses on the quality of the generated criteria.

\subsection{Consistency of LLM-generated Criteria}
Firstly, we explicitly measure the consistency of the criteria generated by the LLM evaluator when given the same task.
We use the prompting template as described in $\S$\ref{sec:prompt_for_criteria} and instruct \texttt{gpt-3.5-turbo} 10 times at a temperature of 0.7, generating multiple criteria sets $\{\mathrm{C}_1,\ldots,\mathrm{C}_{10}\}$.
We also set the temperature as 0 for a deterministic result $\mathrm{\tilde{C}}$ as reference.
We desire that LLMs can perform robustly to generate mostly similar criteria across different samplings and hyperparameters.

Here, we design two embedding-based metrics to estimate the consistency.
\textit{Criteria Consistency} (CC) quantifies the average similarities of matched criteria pairs between the deterministic criteria set $\mathrm{\tilde{C}}$ and a sampled criteria set $\mathrm{C}_n$. 
\textit{Inter-criteria Consistency} (ICC) measures the average similarities of matched criteria pairs between any two sampled set $\mathrm{C}_n$ and $\mathrm{C}_m$.

{\small 
\begin{align}
&\textit{CC}= \frac{\sum_{n=1}^{N}\sum_{i=1}^{|\mathrm{\tilde{C}}|}\max_{\boldsymbol{c}_j \in\mathrm{C}_n} {\textit{sim}(\boldsymbol{\tilde{c}}_i, \boldsymbol{c}_j})}{|\mathrm{\tilde{C}}| \times N}\, ,
\label{eq:cc}\\
&\textit{ICC}= \frac{\sum_{m=1}^{N}\sum_{n=1, m\neq n}^{N}\sum_{i=1}^{|\mathrm{C}_m|}\max_{\boldsymbol{c}_j\in\mathrm{C}_n}\textit{sim}(\boldsymbol{c}_i, \boldsymbol{c}_j)}{\sum_{m=1}^{N}|\mathrm{C}_m| \times (N-1)}\, ,
\label{eq:icc}
\end{align}}

\noindent where $\boldsymbol{c}$ is a singular criterion and $N$ is the number of samplings (10 in our study).
The $\textit{sim}(\cdot)$ represents cosine similarity based on the SimCSE~\citep{gao2021simcse} embeddings of criteria $\boldsymbol{c}_i$ and $\boldsymbol{c}_j$.

\begin{table}[!t]
\centering
\resizebox{0.95\columnwidth}{!}{
\begin{tabular}{llccc}
\toprule
\textbf{Metric} & \textbf{ELI5} & \textbf{ROCStories} & \textbf{GSM8K} & \textbf{Self-Instruct} \\
\midrule
\textit{CC} & 0.75 & 0.82 & 0.80 & 0.78 \\
\textit{ICC} & 0.78 & 0.81 & 0.77 & 0.76  \\
\bottomrule
\end{tabular}
}
\caption{The consistency of the criteria generated by the LLM evaluator, as defined in Equations \ref{eq:cc} and \ref{eq:icc}, across various sampling instances for four distinct domains.} 
\label{tab:criteria_consistency}
\end{table}

\paragraph{Results}
As shown in Table~\ref{tab:criteria_consistency}, we observe a substantial level of consensus in both CC and ICC, especially for story evaluation (ROCStories).
We can safely conclude that LLM can generate consistent criteria for the same task, which may benefit subsequent evaluation stability.

\subsection{Alignment with Human Experts}
\label{sec:criteria-align-with-human}
Next, we examine whether the LLM-generated criteria align with human expertise.
Human experts receive the same prompt\footnote{Compared to what we offer, human experts undoubtedly possess far more knowledge about the task.} as provided in Section~\ref{sec:prompt_for_criteria}.
In our setting, human experts are three researchers with over three years of experience in text generation and language modeling.
Details of human experts are in Appendix~\ref{apx:expert_select}.

\setlength\tabcolsep{5 pt} 
\renewcommand\labelitemi{--} 

\begin{table}[!t]
\centering
\small
\begin{tabular}{p{0.95\linewidth}}
\toprule
\centerline{\textbf{Self-Instruct}}
$\bullet$ {\texttt{$Task$ $Desc.$}}: Give a brief description of the given category of movies and shows. \\
$\bullet$ {\texttt{$Input$}}: Period Dramas \\
$\bullet$ {\texttt{$Output$}}: Want to escape the contemporary world? Explore these historical dramas and shows from the time that have magnificent art and costume design, lots of drama, and a lot of history. \\
\midrule[0.03em]
\centerline{\textbf{Criteria $\mathrm{C}$}} 
\texttt{1.}\hlgreen{\textbf{Coherence}: Does the description flow smoothly and logically?} \cmark \\
\texttt{2.}\hlgreen{\textbf{Accuracy}: Does the description accurately capture the essence of the category of movies and shows?} \bluest{Does it provide a true representation of what viewers can expect from this genre?} \editmark \textcolor{blue}{\textit{remove unknown viewer information}}  \\
\texttt{3.}\redst{Language: Is the language used in the description appropriate and engaging?}  \xmark \textcolor{red}{\textit{unnecessary criterion}} \\
\texttt{4.}\redst{Creativity: Is the description creative and unique?} \xmark \\
\texttt{5.}\redst{Tone: Does the description have an appropriate tone for the category of movies and shows?} \xmark\  \\
\hdashrule[0.5ex]{7.5cm}{1pt}{3mm} \\
\addmark  \ \hlgreen{\textbf{Conciseness}: How brief and concise is the description? Is it easy to understand and comprehend?}\\
\bottomrule
\end{tabular}
\caption{Demonstration of the alignment between a criteria set generated by LLM and the judgments of human experts. 
\cmark\,, \xmark\,, \editmark\,, and \addmark\ denotes the expert's judgments of $Approval$, $Deletion$, $Need\_to\_improve$, and $Missing$, respectively. The criteria agreed by experts are highlighted in \hlgreen{green}.} 
\label{tab:criteria_of_one_instruct}
\end{table}

We assess the degree of alignment between the criteria of LLM and those of human experts from two perspectives: {sufficiency} (whether it is needed for this specific task) and {validity} (whether it is clearly stated and executable during evaluation) as meta-evaluation. Based on the two requirements, we define four levels of (mis)alignments accordingly. An illustrative example is shown in Table~\ref{tab:criteria_of_one_instruct}.

\begin{enumerate}[noitemsep,topsep=0em]
    \item $Approval$: A generated criterion is directly approved by the human expert, e.g., the first criterion in Table~\ref{tab:criteria_of_one_instruct}.
    \item $Need\_to\_improve$: The criterion is necessary, but needs some improvements, including clarifying and making the criterion more executable (e,g, the second criterion requires viewer information that is not available for this task.), and adjusting content to avoid overlap with other criteria.
    \item  $Deletion$: The criterion is unnecessary due to its needless or invalidity. The fourth criterion, ``creativity'' in Table~\ref{tab:criteria_of_one_instruct} is unnecessary when composing a brief description.
    \item $Missing$: A criterion is crucial but missed by the LLM evaluator, e.g., the sixth criterion.
\end{enumerate}

Ideally, a high ratio of $Approval$ is preferred. 
$Need\_to\_improve$ also can express a moderate alignment with human expertise.
$Deletion$ or $Missing$ is not desirable. 

\begin{table}[!t]
\centering
\resizebox{\columnwidth}{!}{
\begin{tabular}{lcccc}
    \toprule
    \textbf{Task} & $Appr.$ & $Need\_to\_Impr.$ &  $Dele.$  & $Miss.$ \\
    \midrule
   ELI5  &  56.52\% & 8.33\% & 35.15\% & ~~~~~0\% \\
   ROCStories & 54.84\% & 9.68\% & 32.26\% & ~~3.23\% \\
   GSM8K & 25.00\%  & ~~~~0\% & 75.00\% & ~~~~~0\%\\
   Self-Instruct &  16.19\% & ~~4.02\% & 79.22\% & ~~0.58\%  \\
    \bottomrule
\end{tabular}
}
\vspace{-1mm}
\caption{Alignment between criteria generated by LLM and those proposed by human experts. The frequent occurrences of 0\% indicate all criteria are either accepted or disapproved by humans.} 
\label{tab:criteria_action_rate}
\vspace{-4mm}
\end{table}

\paragraph{Results}
We compute the ratio of each category as the degree of alignment between the criteria generated by the LLMs and those of human experts\footnote{If there are disagreements among experts, we include a discussion among them and reach a consensus.}. 
As shown in Table~\ref{tab:criteria_action_rate}, the $Approval$ rates across the four benchmarks are considerably high, particularly for ELI5 and ROCStories, where more than 50\% of the criteria proposed by the LLM are accepted by human experts.
The low percentages of $Need\_to\_Improve$ rates on all benchmarks surprise us, even displaying 0\% on two out of the four benchmarks.
This indicates that the LLM can comprehend what one valid criterion should be, and typically does not produce impractical criteria.

The $Deletion$ rates are noteworthy, though, and this is in line with earlier studies that found OpenAI's GPT series to be verbose and repetitious in specific contents~\cite{saito2023verbosity}.
The existence of $Missing$ is not preferred in our results.
We found that \texttt{gpt-3.5-turbo} disregards the specified length requirement in the case of four-sentence or five-sentence story generation. 
Likewise, it fails to consider ``completeness'' in tasks like identifying all words that match a given pattern.
More LLM-generated criteria can be found in Appendix~\ref{app:criteria_analysis_exp}.

\begin{figure}
    \centering
    \includegraphics[width=\linewidth]{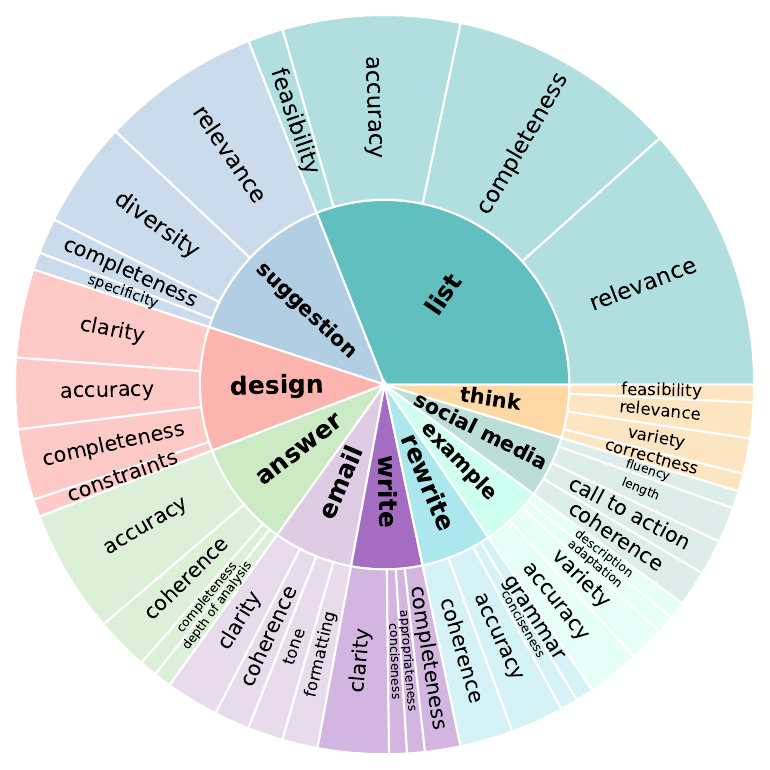}
    \caption{The 10 most frequently occurring key verbs or nouns in the evaluated instruction tasks, as well as the top 4 criteria that are most frequently considered in evaluating the responses to those instructions.}
    \label{fig:self-instruct-criteria-dist}
\end{figure}

\subsection{Criteria Diversity}
Figure~\ref{fig:self-instruct-criteria-dist} visually represents the top 10 most frequent verbs or nouns (out of a total of 96 identified) present in the \texttt{[$task$ $desc.$]} of the Self-Instruct dataset, accompanied by their respective top 4 criteria, which are established by human experts.
By evaluating the criteria generated by LLMs on instruction-following benchmark that encompasses diverse intents, we can gain a comprehensive understanding of their behavior when serving as customized evaluators.

\section{LLMs for Sample-wise Evaluation}
After constructing the high-quality evaluation criteria for three benchmarks and 252 instruction-following tasks, we proceed to examine the reliability of \texttt{gpt-3.5-turbo} as an evaluator based on specific criteria for evaluating a task sample.
For each criterion, we request that the LLM produce an explanation in addition to a score, allowing it to reflect on its own scoring process. 
At last, we prompt the LLM to consider all criteria and their respective scores before determining the overall score.
This procedure allows us to obtain a detailed understanding of the performance of LLMs' tailored evaluators.

\label{sec:instance_eval}
\subsection{Criterion-level Evaluation Prompt}
\label{sec:llm_se}
We adopt a step-by-step instruction, which asks the LLM to evaluate a task sample ($\boldsymbol{x}$, $\boldsymbol{y}$) by considering one criterion $\boldsymbol{c}_i$ at a time.
For each criterion, the LLM is instructed to provide an evaluation with detailed explanations, i.e., (1) a detailed reasoning process to explain the evaluation judgment and (2) a rating score on a 5-point Likert scale\footnote{For the majority of criteria, a maximum score of 5 is adopted. However, for certain non-language-level criteria, such as \textit{length requirement} or \textit{reasoning completeness}, scores are assigned using a 3-level scale.}:
\setlength\tabcolsep{5 pt} 
\renewcommand\labelitemi{--} 

\begin{table}[!t]
\centering
\small
\begin{tabular}{p{0.95\linewidth}}
\toprule
\centerline{\textbf{ELI5}} 
$\bullet$ {\texttt{$Task$ $Desc.$}}: ELI5 is a task for long-form question answering. It contains complex, diverse questions that require explanatory multi-sentence answers. This task aims to provide an explanatory answer that is comprehensible to five-year-olds. \\
$\bullet$ {\texttt{$Input$}}: How is perfume created?    \\
$\bullet$ {\texttt{$Output$}}: Smelly thinks in flowers and herbs can be extracted with alcohol. Then they can be condensed, then put in a bottle, then sprayed on girls and boys alike.   \\
$\bullet$ \texttt{$Criterion$}: Use simple and easy-to-understand language. \\
\midrule[0.03em]
\centerline{\textbf{LLM evaluating on one criterion in each step}}
\hlgreen{1. The answer does not satisfy the criterion as it uses words like ``condensed'' which may not be familiar to a five-year-old.} \cmark \\
\ - \bluest{The use of the words ``condensed'' and ``smelly thinks in flowers and herbs.''} \editmark \ \textcolor{blue}{\textit{wrong explanation}} \hlgreen{- The use of the words ``alcohol'' and ``condensed''.}  \\
\ \hlgreen{- However, the answer does use simple language to explain that perfume is made by extracting scents from flowers and herbs and then putting the condensed scents in a bottle to be sprayed on people.} \cmark \\
2. Score: \bluest{4.} \editmark \ \textcolor{blue}{\textit{The score exceeds the expected value.}} \hlgreen{3.} \\
\bottomrule
\end{tabular}
\vspace{-1mm}
\caption{Illustration of the alignment between LLM's evaluation and human annotators for the ELI5 task~\citep{eli5_lfqa}. Text highlighted in \hlgreen{green} represents the evaluation by human annotators.}
\vspace{-4mm}
\label{tab:evaluation_of_eli5}
\end{table}

\begin{quote}
\small
    \textit{Now, we have a task \texttt{[$task$ $desc.$]}.}
    
    \textit{You need to evaluate whether the output of this task satisfies a given criterion.}

    \texttt{Input: [$\boldsymbol{x}$]}
    \texttt{Output: [$\boldsymbol{y}$]}
    
    \texttt{Criterion: \texttt{[$\boldsymbol{c}_i$]}}

    \textit{Evaluation Steps:}
    \begin{enumerate}[itemsep=\sep, topsep=0.1pt]
    \item \textit{Verify whether the output satisfies the requirement of the given criterion and provide explanations regarding your evaluation.}
    \item \textit{Assign a score to represent your evaluation result on a scale of \texttt{[$Lowest$ $Score$]} to \texttt{[$Highest$ $Score$]}, where \texttt{[$Lowest$ $Score$]} is the lowest and \texttt{[$Highest$ $Score$]} is the highest based on the criterion.}
    \end{enumerate}
    \textit{Evaluation Form:} 
\end{quote}

Providing a detailed explanation before reaching a final evaluation result allows for an in-depth examination of the trustworthiness of LLM's evaluations.
Upon completion of the evaluation of all criteria, the LLM evaluator is obligated to assign an overall quality score by considering all criteria.

\begin{figure}
    \centering
    \includegraphics[width=0.9\linewidth]{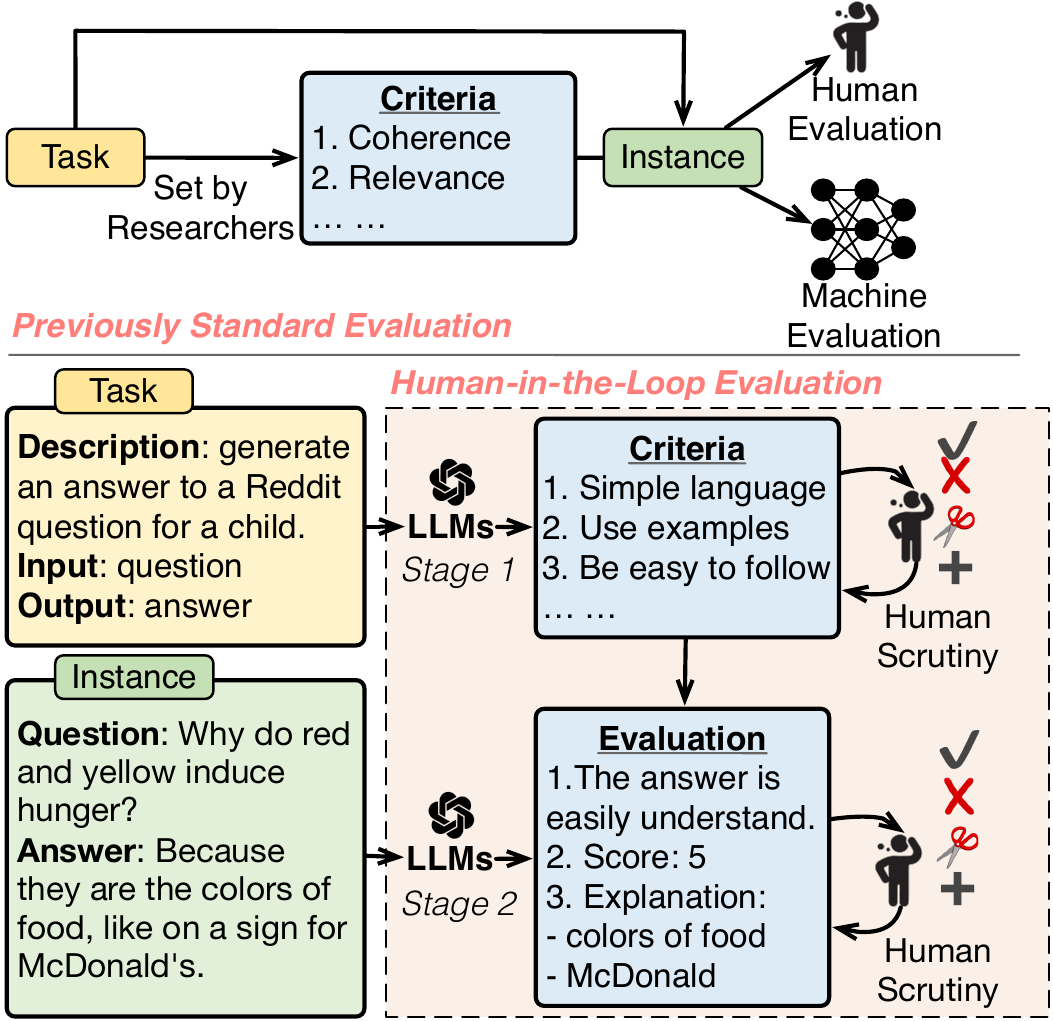}
    \caption{Compared with conventional evaluation methods, the proposed collaborative evaluation pipeline employs an LLM-ideation-human-scrutiny pipeline from task criteria establishment to instance-level evaluation.}
    \label{fig:humanintheloop}
\end{figure}

Besides the criterion-level prompting, we also include a vanilla prompt for the LLM evaluator to directly evaluate the task sample at the overall level.
The detailed prompting format is provided in the Appendix~\ref{app:prompt_template}.

\paragraph{Human-in-the-loop} We also explore a human-in-the-loop setting, where an LLM evaluator assists human evaluators by providing an initial evaluation reference for human annotators to conclude a final judgment. 
We present the human-in-the-loop evaluation pipeline in Figure~\ref{fig:humanintheloop}, which first generates a checklist of task-specific criteria and subsequently conducts instance evaluation. Both stages involve the collaboration between LLM and humans. That is, regarding LLM evaluation results as initial drafts, humans can perform four distinct actions as described in Section~\ref{sec:criteria-align-with-human}, to reach the final evaluation.
The LLM is employed as an assistant for providing diverse criteria and informative evaluations as preliminary references, and then human evaluators scrutinize and make necessary corrections to the outcomes of LLMs, ensuring reliable evaluation while reducing human effort.

\paragraph{Convention Human Evaluation}
In contrast to the LLM evaluators, we ask five professional human annotators to score samples from the three benchmarks (ELI5, ROCStories, and GSM8K) based on the same set of criteria, followed by an overall score.
For Self-Instruct, it is very difficult to find satisfactory human annotators because different tasks require different expertise, so we directly use the released expert annotation statistics from~\citet{wang2022self}.

\subsection{Weaknesses of LLM evaluators}
\label{exp:llm-evaluator}
To compare the disparity between the LLM evaluator and human annotators, we first compute their Pearson and Spearman correlation.
Next, we examine the scoring distribution, including the distribution shift with human annotations and the scoring bias on LLM's evaluation output.

\begin{table}[!t]
\centering
\tiny
\resizebox{\columnwidth}{!}{
\begin{NiceTabular}{l|cc|cc|cc}  
\toprule
\multirow{2}{*}{\textbf{Prompt}}& \multicolumn{2}{c}{\textbf{ELI5}} & \multicolumn{2}{c}{\textbf{ROCStories}} & \multicolumn{2}{c}{\textbf{GSM8K}}  \\
&  $r$ & $\rho$  & $r$ & $\rho$  & $r$ & $\rho$ \\
\midrule
Direct &  0.128  & 0.142 & 0.199 & 0.217  & $\operatorname{NaN}$  & $\operatorname{NaN}$  \\
\midrule
Step-by-step & 0.407 & 0.392 & 0.282  & 0.200  & -0.016 & -0.018 \\

\midrule
\makecell[l]{Step-by-step\\+Human-in-the-loop} & 0.412 & 0.417 & 0.427 & 0.437 &  0.669 & 0.612 \\
\bottomrule
\end{NiceTabular}
}
\vspace{-1mm}
\caption{Sample-level Pearson ($r$) and Spearman ($\rho$) correlations of the overall scores on three benchmarks.} 
\vspace{-3mm}
\label{tab:overall_eval_correlation}
\end{table}

\begin{figure}[!t]
    \centering
    \includegraphics[width=\linewidth]{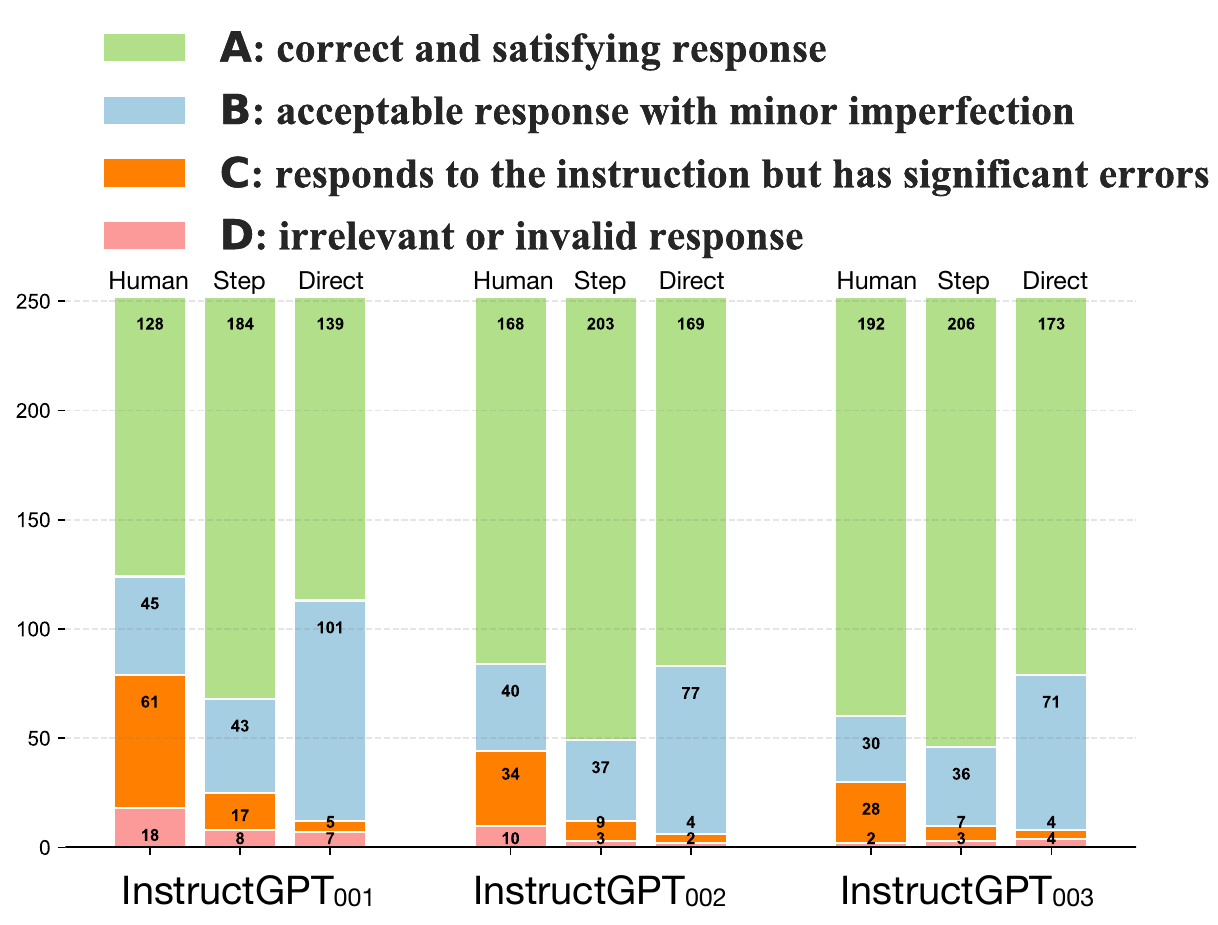}
    \caption{The distribution of overall quality scores for three models, InstructGPT$_{001}$, InstructGPT$_{002}$, and InstructGPT$_{003}$, on the Self-Instruct dataset. These scores are evaluated by human experts (Human), LLM with step-by-step evaluation (Step), and LLM with direct evaluation (Direct), respectively.}
    \label{fig:self_instruct_score_dist}
\end{figure}

\begin{table*}[!t]
\centering
\tiny
\resizebox{1.8\columnwidth}{!}{
\begin{NiceTabular}{l|cc|cc|cc|cc|cc}  
\toprule
\multirow{2}{*}{\textbf{Evaluation}}& \multicolumn{2}{c}{\textbf{Comprehensibility}} & \multicolumn{2}{c}{\textbf{Accuracy}} & \multicolumn{2}{c}{\textbf{Coherence}}  & \multicolumn{2}{c}{\textbf{Engagement}}  & \multicolumn{2}{c}{\textbf{Analogy usage}}  \\
 & $r$ &  $\rho$  & $r$ &  $\rho$  & $r$  & $\rho$  & $r$ & $\rho$  & $r$ & $\rho$ \\
\midrule
Direct & 0.231 & 0.209 & 0.076 & 0.073 & 0.098 & 0.092 & 0.102 & 0.087 & $\operatorname{NaN}$ & $\operatorname{NaN}$  \\
\midrule
Step-by-step &  0.288 & 0.257 & 0.410 & 0.361  & 0.393 & 0.366 & 0.499 &  0.399  & 0.261  & 0.246 \\
\midrule
\makecell[l]{Step-by-Step\\+Human-in-the-loop} & 0.303& 0.310 & 0.410 & 0.413 & 0.491 & 0.499 &  0.429 & 0.432 & 0.396 & 0.317 \\
\bottomrule
\end{NiceTabular}
}
\caption{Sample-level Pearson ($r$) and Spearman ($\rho$) correlations between human annotators and LLM with direct evaluation, as well as LLM with step-by-step evaluation, on the ELI5 dataset for different criteria. In cases where the correlation is ``$\operatorname{NaN}$'', the LLM assigns identical scores to all outputs under that particular criterion.} 
\label{tab:eli5_eval_correlation}
\end{table*}

\begin{table*}[!t]
\centering
\tiny 
\resizebox{1.85\columnwidth}{!}{
\begin{NiceTabular}{l|cc|cc|cc|cc|cc|cc}  
\toprule
\multirow{2}{*}{\textbf{Evaluation}}& \multicolumn{2}{c}{\textbf{Relevance}} & \multicolumn{2}{c}{\textbf{Coherence}} & \multicolumn{2}{c}{\textbf{Language}}  & \multicolumn{2}{c}{\textbf{Commonsense}} & \multicolumn{2}{c}{\textbf{Creativity}}   & \multicolumn{2}{c}{\textbf{Length}} \\
 & $r$ &  $\rho$  & $r$ &  $\rho$  & $r$  & $\rho$  & $r$ & $\rho$  & $r$ & $\rho$ & $r$ & $\rho$ \\
\midrule
Direct & 0.169 &  0.154 & 0.155 & 0.153 & 0.165 & 0.167 & $\operatorname{NaN}$ & $\operatorname{NaN}$ & 0.028  & 0.026  & $\operatorname{NaN}$ & $\operatorname{NaN}$  \\
\midrule
Step-by-Step &  0.245 & 0.241 & 0.237 & 0.231  & 0.256 & 0.271 & 0.266 &  0.265 & 0.130 & 0.113  & 0.118 & 0.118\\
\midrule
\makecell[l]{Step-by-Step\\+Human-in-the-loop} & 0.415 &  0.425 & 0.398 & 0.409 & 0.418 & 0.430 & 0.400 & 0.376 &  0.388 & 0.395  & 0.800 & 0.794 \\
\bottomrule
\end{NiceTabular}
}
\caption{Sample-level Pearson ($r$) and Spearman ($\rho$) correlations of different criteria on the ROCStories dataset.} 
\label{tab:rocstories_eval_correlation}
\end{table*}

\begin{table}[!t]
\centering
\footnotesize 
\resizebox{\columnwidth}{!}{
\begin{NiceTabular}{l|ccc|ccc|ccc}  
\toprule
\multirow{2}{*}{\textbf{Evaluation}}& \multicolumn{3}{c}{\textbf{Logical Reasoning}} & \multicolumn{3}{c}{\textbf{Numerical Understanding}}  & \multicolumn{3}{c}{\textbf{Completeness}} \\
 & \textsc{CS} & \textsc{OE}  & \textsc{ME} &  \textsc{CS} & \textsc{OE}  & \textsc{ME} & \textsc{CS} & \textsc{OE}  & \textsc{ME} \\
\midrule
Direct & \textcolor{blue}{100} & 0 & 0 & \textcolor{blue}{100} & 0 & 0 & \textcolor{blue}{100} & 0 & 0 \\
\midrule
Step-by-step &  \textcolor{blue}{100} & 0 & 0 & 99 & 0 & 1 & \textcolor{blue}{100} & 0 & 0 \\
\bottomrule
\end{NiceTabular}
}
\caption{Distribution of scores between evaluations generated by LLM and human evaluators on GSM8K dataset. The \textcolor{blue}{100} indicates that the evaluator considers all solutions correct for the corresponding criterion. \textsc{CS} represents a ``correct solution'', \textsc{OE} indicates ``one error exists'', and \textsc{ME} means ``multiple errors exist''.} 
\label{tab:gsm8k_eval_correlation}
\end{table}

\paragraph{Correlations between LLM-based Scoring and Human Judgments}
Table~\ref{tab:overall_eval_correlation} demonstrates the correlation between human evaluations and various evaluation techniques (i.e., vanilla prompting, step-by-step prompting, and the integration of LLMs and human involvement) across the three benchmark datasets.
Prompting LLMs without considering task-specific criteria yields poor performance on all three datasets. 
In comparison, the step-by-step prompting significantly improves human correlation, particularly on the ELI5 dataset. Nonetheless, when compared to the human-in-the-loop setup, the LLM evaluation exhibits lower correlations on the ROCStories dataset (which entails creativity and subjectivity) and the GSM8K dataset (which requires mathematical reasoning).

Figure~\ref{fig:self_instruct_score_dist} illustrates the comparison between LLM-based evaluation and human evaluation for the Self-Instruct task.
Compared to pure human evaluation and LLM evaluation with step-by-step criterion evaluation, the LLM evaluator without specific criteria cannot differentiate between ``minor imperfections'' and ``significant errors''.
Step-by-step evaluation along each criterion can detect severe errors but tends to assign higher scores than human evaluators.

The correlations between the LLM evaluator and human annotations on the ELI5 and ROCStories tasks are presented in Table~\ref{tab:eli5_eval_correlation} and Table~\ref{tab:rocstories_eval_correlation} respectively.
The LLM evaluator performs well on general language-level criteria (e.g., \textit{relevance} and \textit{coherence}) but poorly on criteria involving information seeking (e.g., analogy usage, creativity) or numerical judgment (e.g., \textit{length}), showing weak correlations with humans.

\paragraph{Scoring Distribution Bias}

Considering the unsatisfactory correlation with human annotators, we aim to investigate the disparity between the sample-wise scoring distribution of LLM-based evaluation and human evaluation. 
Firstly, we present the LLM evaluator's performance on the GSM8K dataset in Table~\ref{tab:gsm8k_eval_correlation}, where significant difficulties are encountered.
Surprisingly, the LLM evaluator consistently assigns the highest scores to samples, regardless of the presence of errors in logic, numbers, or completeness, indicating the significant challenges faced by LLMs in detecting math-related errors.

\begin{figure}[!t]
    \centering
    \includegraphics[width=0.9\linewidth]{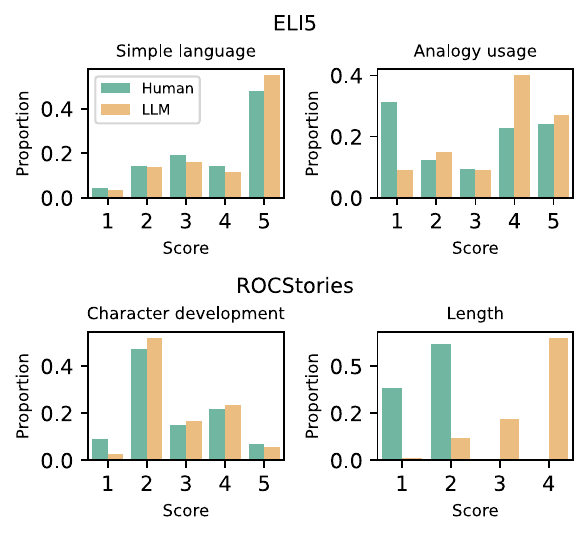}
    \vspace{-0.3cm}
    \caption{The distribution of scores assigned by LLM and 5 human evaluators for predictions generated by various models. To ensure generalization, human evaluators who participated in different datasets may vary.}
    \label{fig:score_correction_pattern}
\end{figure}

For the question-answering task (ELI5) and story generation task (ROCStories), the distribution of scores assigned by both LLM and human evaluators is presented in Figure~\ref{fig:score_correction_pattern}. 
To facilitate comparison, we select two representative criteria for each task. 
For the criteria on the left side, the difference in scoring distribution between LLM and humans is negligible, demonstrating the effectiveness of LLM evaluation on these criteria.
However, for criteria on the right side, which pertain to information-seeking and numerical capabilities, there is a significant difference. The difference is large, indicating LLM evaluation still may fail on those complex evaluation criteria.

\begin{table}[!t]
\centering
\tiny
\resizebox{\columnwidth}{!}{
\begin{NiceTabular}{l|c|c|c|c}  
\toprule
\textbf{Evaluator} & \textbf{Score=1} & \textbf{Score=2} & \textbf{Score=3} & \textbf{Score=4} \\
\midrule
GPT-3.5-turbo & 3.17 &	24.21	&17.86&	54.76 \\
\midrule
GPT-4 & 8.77	&19.30&	29.82	&42.11 \\

\midrule
HumanEval & 2.77	&1.98	&40.08&	55.17 \\
\bottomrule
\end{NiceTabular}
}
\caption{Evaluation score distribution of LLM evaluators and human evaluators.} 
\label{tab:gpt4_result}
\end{table}

\paragraph{Evaluating with GPT-4}

Our analysis of the more powerful model, \texttt{gpt-4}, as an evaluator, displayed close alignment with the performance of \texttt{gpt-3.5-turbo}. The results are shown in Table~\ref{tab:gpt4_result}. 
LLMs showed a consistently positive and negative tendency akin to humans, as observed in prior research~\citep{zheng2023judging, wu2023style}, but tended to provide moderate scores in comparison. To address this, we recommend starting with sample-wise evaluations before progressing to pair-wise evaluations, a more reliable strategy supported by our study findings.

\subsection{Can We Enhance the Evaluation with LLM-human-in-the-loop?} 
\label{exp:llm_human_eval}
From the above results and analysis, we can see that the current LLM evaluator (\texttt{gpt-turbo-3.5} in this instance) is still imperfect. 
A cost-effective and reliable approach to enhance evaluation is to involve LLMs as auxiliary evaluators for assisting human evaluators.

\paragraph{Correlation with Convention Human Evaluation}
To fairly investigate the effectiveness of LLM-human-in-the-loop, we calculate the average Pearson correlation among pairs of evaluators for each task in Table~\ref{tab:compare_iaa_with_pure_human}.
We can observe that, with the LLM serving as an assistant, the correlation between human-in-the-loop evaluation and pure human evaluation is similar to the internal correlation among pure human annotators. This finding implies that humans are not significantly influenced by the biases of the LLM.
Nonetheless, the scores of LLM evaluation exhibit a relatively weak correlation with those of humans, suggesting that relying solely on LLM evaluation may not be reliable.

\begin{table}[!t]
\setlength{\belowcaptionskip}{-0.2cm}
\centering
\resizebox{0.9\columnwidth}{!}{
\begin{tabular}{lccc}
    \toprule
    \textbf{Task}  & \textbf{LLM vs. \linebreak \textsc{HuE}} & \textbf{\textsc{L}+\textsc{H} vs. \textsc{HuE}} & \textbf{\textsc{HuE}} \\
    \midrule
    ELI5 & 0.21 & 0.31 & 0.40 \\
    ROCStories & 0.35 & 0.43 & 0.45 \\
    Self-Instruct & 0.37 & 0.33 & 0.23  \\
    \bottomrule
\end{tabular}
}
\caption{Average Pearson correlation among pairs of evaluators with pure human evaluation, i.e., LLM, \textsc{LLM+HumanEval} (Human-in-the-loop evaluation), and \textsc{HumanEval}.}  
\label{tab:compare_iaa_with_pure_human}
\end{table}

\begin{figure}[t!]
    \centering
    \includegraphics[width=0.8\linewidth]{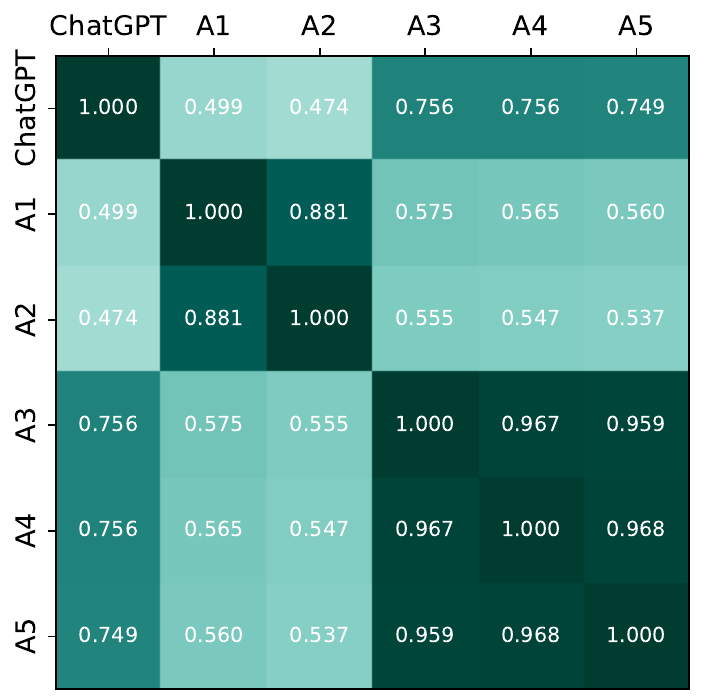}
    \caption{Inter-annotator agreement among LLM and 5 humans (A1 to A5) using Krippendorff's $\alpha$. LLM's scores are deemed acceptable, with over 50\% of human evaluators showing high agreement ($\alpha > 0.7$).}
    \label{fig:consistency}
\end{figure}

\paragraph{Inter Annotator Agreements of LLM-evaluation with Human-in-the-loop}

\label{sec:step-by-step prompting with human-in-the-loop}
Figure~\ref{fig:consistency} reports inter-annotator agreement (IAA) when adopting LLM-evaluation with human-in-the-loop setting on 200 randomly-sampled ELI5 samples.
We use Krippendorff's $\alpha$ for evaluation scores to showcase the annotation consistency among LLM and 5 involved human evaluators.
Our findings reveal high agreement among evaluators when evaluating samples based on the initial drafts proposed by the LLM evaluator, with $\alpha$ exceeding 0.8 for two groups of evaluators, i.e., (Group 1: A3, A4, and A5) and (Group 2: A1 and A2), indicating inherent variance in the definitions of ``best text snippets'' among different evaluators. Notably, Group 1 (with $\alpha \approx 0.754$) is more consistent with LLM's evaluation scores compared to Group 2 (with $\alpha \approx 0.487$). 

\begin{table}[!t]
\setlength{\belowcaptionskip}{-0.3cm}
\centering
\resizebox{\columnwidth}{!}{
\begin{tabular}{lcccc}
    \toprule
    \textbf{Task}  & \textbf{Correction} (\%) & \textbf{Scrutiny} (\%) & \textbf{Subjectivity} (\%) & \textbf{Outlier} (\%) \\
    \midrule
    ELI5 & $\text{42.55}_{47}$ & $\text{55.32}_{47}$  & $\text{76.47}_{17}$ & $\text{63.64}_{22}$ \\
    ROCStories & $\text{11.43}_{35}$  & $\text{14.29}_{35}$ & $\text{37.50}_{24}$  & $\text{72.73}_{33}$   \\
    Self-Instruct & $\text{45.45}_{11}$  & $\text{45.45}_{11}$ & $\text{61.90}_{21}$  & $\text{48.39}_{31}$  \\
    \bottomrule
\end{tabular}
}
\vspace{-0.2cm}
\caption{Human evaluator behaviors in LLM-Human-in-the-loop evaluation. 
\textbf{Correction}: humans adjust LLM's conflicting evaluations to align with \textsc{HumanEval}
\textbf{Scrutiny}: humans scrutinize LLM evaluations inconsistent with \textsc{HumanEval}.
\textbf{Subjectivity}: disagreements arise among human evaluators when \textsc{HumanEval} shows large disparities.
\textbf{Outlier}: all participants agree with the majority vote of \textsc{HumanEval}, despite some disagreements.
The subscript number indicates the corresponding instances of \textsc{HumanEval}.}  
\label{tab:finegrained_human}
\end{table}

\paragraph{Reasons behind the Relatively High Correlation of the Human-in-the-loop Evaluation}
Given the improved inter-annotator agreement of step-by-step prompting with human-in-the-loop, we carefully investigate the factors that improve human agreements. 
We choose the majority vote of human annotators as ground truth to analyze the behavior of humans involved in step-by-step prompting with human-in-the-loop. The results are presented in Table~\ref{tab:finegrained_human}.
The elevated values observed in the ``Correction'', ``Scrutiny'', and ``Subjectivity'' categories suggest that humans tend to follow their own preferences in most cases. This is evident from their endeavors to revise LLM's evaluations that contradict their own judgments (55.32\% on ELI5) without blindly relying on LLM (61.90\% on Self-Instruct).
Surprisingly, the notable values in the ``Outlier'' indicate that humans are willing to agree with LLM when it is justified and reasonable.
This willingness to align with LLM evaluations leads to higher annotator agreements and removes the outliers existing in human annotators.

\section{Conclusion}
To examine the reliability of LLMs as universal evaluators, we investigate whether LLMs can generate appropriate evaluation criteria across various tasks and whether the evaluation results are trustworthy based on the given criteria. We request LLMs to create a draft evaluation, then human experts are employed to assess and refine the draft evaluation. 
Based on the expert assessment, we find that 1) LLMs can consistently generate high-quality task-specific evaluation criteria, while also producing many unnecessary criteria or missing a few crucial criteria. 2) LLMs perform well on language-level and commonsense-related criteria while making mistakes on complex criteria, such as ``analogy usage'' and ``logical reasonability''. 
After introducing human refinement, the LLM evaluator can mitigate specific human subjectivity with reduced annotation outliers.

\section*{Limitations}

We use \texttt{gpt-turbo-3.5} as our specific LLM evaluator to assess the reliability of the LLM-as-judge paradigm due to its cost-effectiveness and performance.
Given the high cost of human evaluation, we did not include additional LLMs. Hiring a qualified evaluator for 200 instances costs us \$700 and takes over three weeks to recruit, test, and collect results, while guiding \texttt{gpt-turbo-3.5} for evaluations takes just a few hours.

LLM evaluation results can be sensitive to instruction formats. While finding a globally optimal instruction is challenging, our pilot experiments show that results across different instructions are not significantly different for a small subset of instances.
Additionally, decoding strategies influence evaluation outcomes. For deterministic results, we set the sampling temperature to 0, while also experimenting with a temperature of 0.7 and sampling 10 times. We observe that this variation had no significant impact on LLM evaluations.

Different from previous work, we are not trying to explore the use of LLMs as evaluators for any new particular task or to design any better prompt for the LLM to fulfill the task as evaluators.

\section*{Ethics Statement}
We honor the Code of Ethics. 
No private data or non-public information is used in this work.
For human evaluation, we recruited our evaluators from the linguistics departments of local universities through public advertisement with a specified pay rate. 
All of our evaluators are senior undergraduate students or graduate students in linguistic majors who took this evaluation as a part-time job. 
We pay them US\$35.5 an hour. The local minimum salary in the year 2023 is US\$15.5 per hour for part-time jobs. The annotation does not involve any personally sensitive information.

We use the models and datasets by their intended usage. Specifically, we follow the \href{https://openai.com/policies/usage-policies}{OpenAI usage policy} when using the models of Open AI.

\normalem
\bibliography{custom}

\appendix

\newpage

\section{Evaluation Prompts}
\label{app:prompt_template}

\paragraph{Preparing Evaluation Target Samples}
We carefully designed the query formats for each dataset to guide the models to behave according to the task requirements\footnote{We adopted the prompt design from \url{https://github.com/bigscience-workshop/promptsource} as a reference.}.

\paragraph{Step-by-step Prompt for the Overall Evaluation}
After the evaluations on the criteria level are generated, we assess the LLM evaluator's capability to assign overall quality scores comparable to those given by human annotators, taking into account all the criteria.
In this analysis, the previous criterion-level evaluation is treated as a multi-turn history, and an overall score is generated based on multiple perspectives. 

\begin{quote}
    \textit{Now, we have a task \texttt{[$task$ $desc.$]}.}
    
    \texttt{Input: [$\boldsymbol{x}$]}
    \texttt{Output: [$\boldsymbol{y}$]}
    
    [The previous evaluation for each criterion is omitted here.]

    \textit{Based on the provided input and evaluation of multiple criteria, you need to evaluate the overall quality of the output for this task.}

    \textit{Evaluation Steps: }

    \begin{enumerate}[itemsep=\sep, topsep=0.1pt]
    \item \textit{Verify the overall quality of output and provide explanations regarding your evaluation.}
    \item 
    \textit{Please an overall score on a scale of \texttt{[$Lowest$ $Score$]} to \texttt{[$Highest$ $Score$]} to represent the quality of the output, where \texttt{[$Lowest$ $Score$]} is the lowest and \texttt{[$Highest$ $Score$]} is the highest based on the criterion. Please make sure you remember the task, the input and output to be evaluated, the multiple criteria, and the corresponding scores you assigned.}
    \end{enumerate}

    \textit{Evaluation Form:}
\end{quote}

\paragraph{Straightfoward Prompt for the Overall Evaluation}
In contrast to the criterion-level evaluation approach, we also include a straightforward prompt to enable the LLM to directly generate overall scores without prior evaluation of specific criteria. 

\begin{quote}
    \textit{Now, we have a task \texttt{[$task$ $desc.$]}.}

    \textit{You need to evaluate the overall quality of the output of this task.}

    \texttt{Input: [$\boldsymbol{x}$]}
    \texttt{Output: [$\boldsymbol{y}$]}

    \textit{You are required to assign an overall score on a scale of 1 to 5 to represent the quality of the output, where 1 is the lowest and 5 is the highest based on the criterion.}

    \textit{Evaluation Score:}
\end{quote}

\section{Can LLM Generate Sufficient Specialized Evaluation Criteria?}
\label{app:criteria_analysis_exp}

Human experts are involved in the evaluation of the criteria generated by LLM. We invite seven NLP researchers with over three years of research experience in text generation and language modeling.
\label{apx:criteria_derivation_roc_self_instruct}
Table~\ref{app:criteria_details} presents the finalized evaluation criteria for three benchmark tasks, while Table~\ref{exp:eval_detail_self_instruct} outlines the criteria to evaluate several example tasks in Self-Instruct.
All finalized criteria have been unanimously agreed upon by experts.

\setlength\tabcolsep{5 pt} 
\renewcommand\labelitemi{--} 

\begin{table}[!t]
\centering
\small
\begin{tabular}{p{0.95\linewidth}}
\toprule
\centerline{\textbf{ELI5}}
$\bullet$ {\texttt{$Task$ $Desc.$}}: Provide an answer to a Reddit forum question that is comprehensible by five-year-olds.    \\
$\bullet$ {\texttt{$Input$}}: What is happening in my mouth when I whistle?  \\
$\bullet$ {\texttt{$Output$}}: You're pushing air past your lips, and the shape of your lips is vibrating the air as it passes by (similar to how your vocal chords vibrating make the sound of your voice on the air). When air is forced through a constricted area, it will vibrate when it hits a surface (which is why when it's windy outside, you can hear the wind whistling through the attic or some other enclosed space). \\
\midrule[0.03em]
\textbf{Criteria $C$:} \\
\ \texttt{1.} \hlgreen{Use simple and easy-to-understand language.} [\cmark] \\
\ \texttt{2.} \hlgreen{Use examples or analogies that are relatable to a five-year-old's experience.}  [\cmark] \\
\ \texttt{3.} \redst{Avoid using technical terms or jargon.}  \xmark \textcolor{red}{\textit{same as the first criterion}} \\
\ \texttt{4.} \redst{Break down complex ideas into smaller, more manageable parts.} \xmark \textcolor{red}{\textit{same as the first criterion}} \\
\ \texttt{5.} \redst{Use visual aids or illustrations to help explain the answer.}   \xmark \textcolor{red}{\textit{not consider visual modality}} \\
\ \texttt{6.} \bluest{Be helpful and understand the child's level of comprehension.} \editmark \textcolor{blue}{\textit{we only consider the overall performance}} \hlgreen{All things considered, answers should be helpful to the person who asked this question.}  \\
\hdashrule[0.5ex]{7.5cm}{1pt}{3mm} \\
\addmark  \ \hlgreen{Answers should be factually correct and cannot have subtly incorrect or fabricated information.} \\
\addmark  \ \hlgreen{Be easy to follow and logically coherent.}  \\
\bottomrule
\end{tabular}
\caption{Demonstration of the derivation of evaluation criteria for the long-form question-answering task ELI5, achieved through the collaboration of LLM ideation and human evaluator correction (comments in square brackets).
\cmark\ signifies approve ($a_{\mathrm{apv}}$), \xmark\ indicates delete ($a_{\mathrm{del}}$), \editmark\ denotes revise ($a_{\mathrm{revise}}$), and \addmark\ represents add ($a_{\mathrm{add}}$). The ultimate criteria are highlighted in \hlgreen{yellow}.}
\label{tab:criteria_of_eli5}
\end{table}

\setlength\tabcolsep{5 pt} 
\renewcommand\labelitemi{--} 

\begin{table}[!t]
\centering
\small
\begin{tabular}{p{0.95\linewidth}}
\toprule
\centerline{\textbf{ROCStories}}
$\bullet$ {\texttt{$Task$ $Desc.$}}: ROCStories is a task for commonsense short story generation. The task aims to generate stories that contain a variety of commonsense causal and temporal relations between everyday events. \\
$\bullet$ {\texttt{$Input$}}: Write a five-sentence story about an everyday topic ``pizza night'' \\
$\bullet$ {\texttt{$Output$}}:  Ann and her mom had a girls' night. They watched movies all night. Then they got hungry. They decided to order a pizza. Girls' night became pizza night! \\
\midrule[0.03em]
\centerline{\textbf{Criteria $\mathrm{C}$}} 
\texttt{1.} \hlgreen{\textbf{Relevance}: be relevant to the given prompt or topic.}  \cmark \\
\texttt{3.} Coherence: \bluest{have a logical flow and provide a closure that makes sense to the reader.}  \editmark \textcolor{blue}{\textit{remove unknown reader information}}
\hlgreen{have a logical flow with a closure.} \\
\texttt{4.} \hlgreen{\textbf{Length}: be an appropriate length for the given task.} \cmark \\
\texttt{5.} \hlgreen{\textbf{Engagement}: be engaging from beginning to end.} \cmark \\
\texttt{7.} \redst{Language: The language should be appropriate for the target audience.} \xmark\ \textcolor{red}{\textit{unnecessary criterion}} \\
\texttt{8.} \redst{Creativity: be creative and unique.} \xmark\ \textcolor{red}{\textit{unnecessary criterion}} \\
\bottomrule
\end{tabular}
\caption{Demonstration of the alignment between a criteria set generated by LLM and the judgments of human experts.
\cmark\,, \xmark\,, \editmark\,, and \addmark\ denotes the expert's judgments of $Approval$, $Deletion$, $Need\_to\_improve$, and $Missing$, respectively. The criteria agreed by experts are highlighted in \hlgreen{green}.} 
\label{tab:criteria_of_roc}
\end{table}

As discussed in Section~\ref{sec:criteria_gen}, we present the experimental results on the consistency of the criteria generated by the LLM and its alignment with human expertise, illustrated through the example task of Self-Instruct (see Table~\ref{tab:criteria_of_one_instruct}).
The comprehensive criteria generated by the LLM and the evaluation results from human experts for ELI5 and ROCStories are presented in Tables~\ref{tab:criteria_of_eli5} and \ref{tab:criteria_of_roc}, respectively.
We do not include full results of Self-Instruct samples, since one criteria set is provided for each instruction and the complete results for all instructions are somewhat long.

\begin{table}[!t]
\setlength{\belowcaptionskip}{-0.2cm}
\centering
\resizebox{\columnwidth}{!}{
\begin{tabular}{lcccc}
    \toprule
    \textbf{Task} & $Approval$ & $Need\_to\_improve$ &  $Deletion$  & $Missing$ \\
    \midrule
   ELI5 & 81.16\% & 14.61\%  & 3.60\% & 0.63\% \\
   -scr. & 70.66\% & 29.34\% & ~~~~~0\% & ~~~~~0\% \\
   -evd. & 85.01\% & 7.17\% & 6.66\% & 1.16\%\\
   \midrule
   ROCStories  & 94.65\% &  3.92\% & 1.09\% & 0.34\%  \\ 
   -scr. & 85.87\% & 14.13\% & ~~~~~0\% & ~~~~~0\% \\
   -evd. & 84.87\% & 13.43\% & 1.29\% & 0.41\%\\
   \midrule
   Self-Instruct & 91.49\% & 4.99\% & 2.95\% & 0.57\% \\ 
   -scr. & 85.06\% & 14.94\% & ~~~~~0\% & ~~~~~0\% \\
   -evd. & 92.88\% & 1.84\% & 4.43\% & 0.85\% \\
    \bottomrule
\end{tabular}
}
\caption{We present the correction rates of human annotators on LLM evaluation results, along with a fine-grained analysis of human preferences regarding the components of the evaluation results, namely conclusion, score, and evidence.}  
\label{tab:eval_action_rate}
\end{table}

\begin{figure}[!t]
    \centering
    \includegraphics[width=0.9\linewidth]{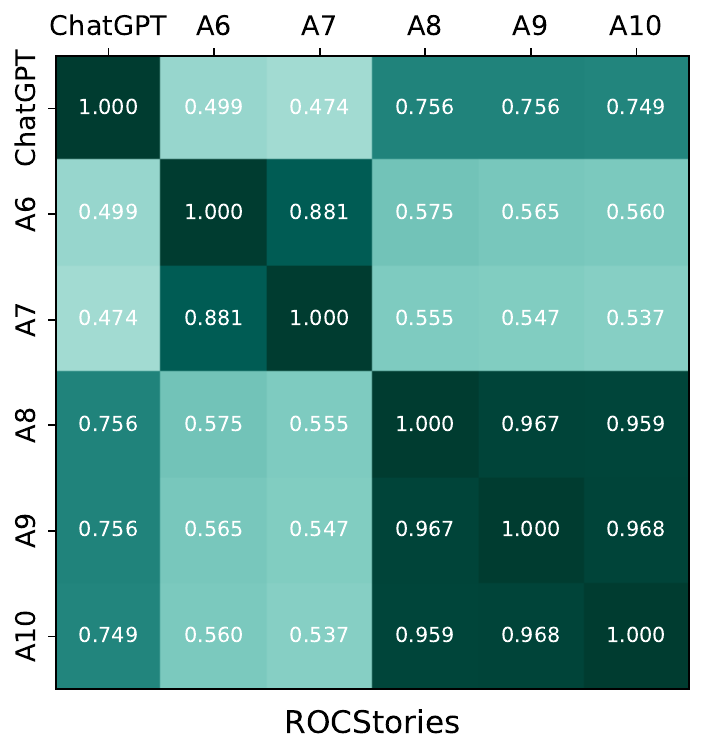}
    \caption{Inter-annotator agreement (ROCStories) among ChatGPT and humans using Krippendorff's $\alpha$. ChatGPT's evaluation scores are deemed acceptable, with over 50\% of human evaluators showing high agreement ($\alpha > 0.7$).}
    \label{fig:roc_consistency}
\end{figure}

\begin{figure}[!t]
    \centering
    \includegraphics[width=0.9\linewidth]{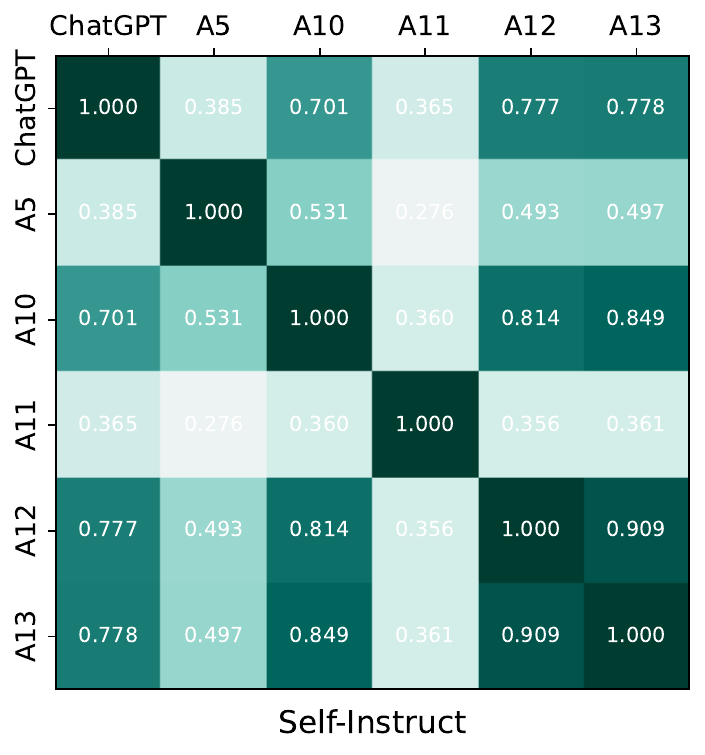}
    \caption{Inter-annotator agreement (Self-Instruct) among ChatGPT and humans using Krippendorff's $\alpha$. ChatGPT's evaluation scores are deemed acceptable, with over 50\% of human evaluators showing high agreement ($\alpha > 0.7$).}
    \label{fig:self_consistency}
\end{figure}

\section{Can LLM Generate Sample-wise Evaluations Aligned with Human Judgments?}

\paragraph{Extraction for LLM's Evaluation Scores}
We observe that LLM tends to provide its evaluation scores in various expressions. 
To extract the scores, we apply three simple rules:
(1) We remove string ``2.'' from the output
since the evaluation score is behind the evaluation conclusion, LLMs will sometimes say ``2. The score \ldots''. 
(2) We remove the string
``out of 5'' and ``/5'' since LLMs sometimes say "give a score of x out of 5'' or ``x/5''
(3) We use the regular expression to extract the first number in the sequence.

\paragraph{Evaluation Consistency Between LLM and Humans}
We initially conducted preliminary experiments to evaluate the overall quality of LLM's step-by-step evaluation outcomes based on given criteria. This is performed using a four-level alignment estimation, as presented in Table~\ref{tab:eval_action_rate}.
When aggregating the preferences of all human evaluators, we observe that the ratios of $Approval$ are consistently high for both the overall evaluations across all datasets, indicating the huge potential of LLM in evaluation. 
It is worth noting that 29.34\% of LLM-generated scores are revised by human evaluators for ELI5. This demonstrates that human involvement is essential in identifying issues overlooked by LLMs.

\paragraph{Evaluation Consistency Among Human Evaluators}
We calculate the inter-annotator agreements of ROCStories and Self-Instruct tasks and present the results in Figure~\ref{fig:roc_consistency} and Figure~\ref{fig:self_consistency} accordingly.
Although the consistencies between evaluator A11 and other evaluators in Figure~\ref{fig:self_consistency} are not very high, the agreements among the other four evaluators are higher than the reliable threshold 0.677.

\begin{figure}[!t]
    \centering
    \includegraphics[width=0.9\linewidth]{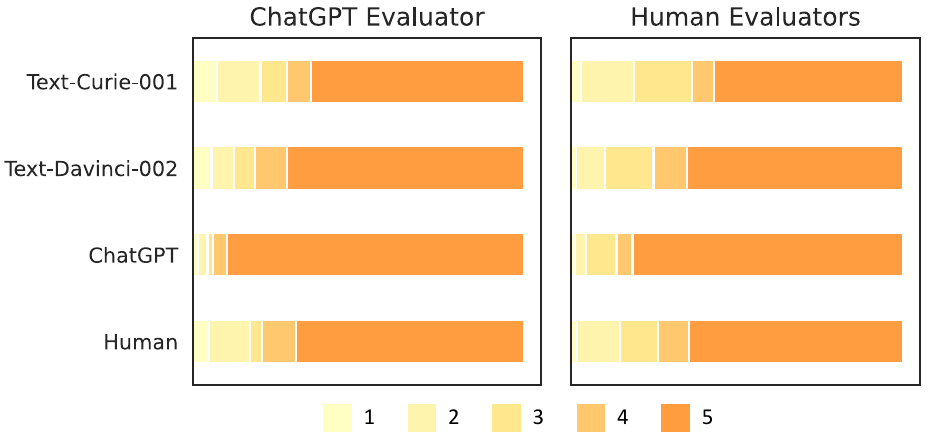}
    \caption{The distribution of scores assigned by LLM and human evaluators (integrated from a group of five individuals) for generations written by humans and models with different qualities on the Self-Instruct task. The score (1 to 5) ratios across different sources are distinct, suggesting that both LLMs and humans can discern these differences.}
    \label{fig:score_of_diff_qualities}
\end{figure}

\begin{figure}[!t]
    \centering
    \includegraphics[width=0.9\linewidth]{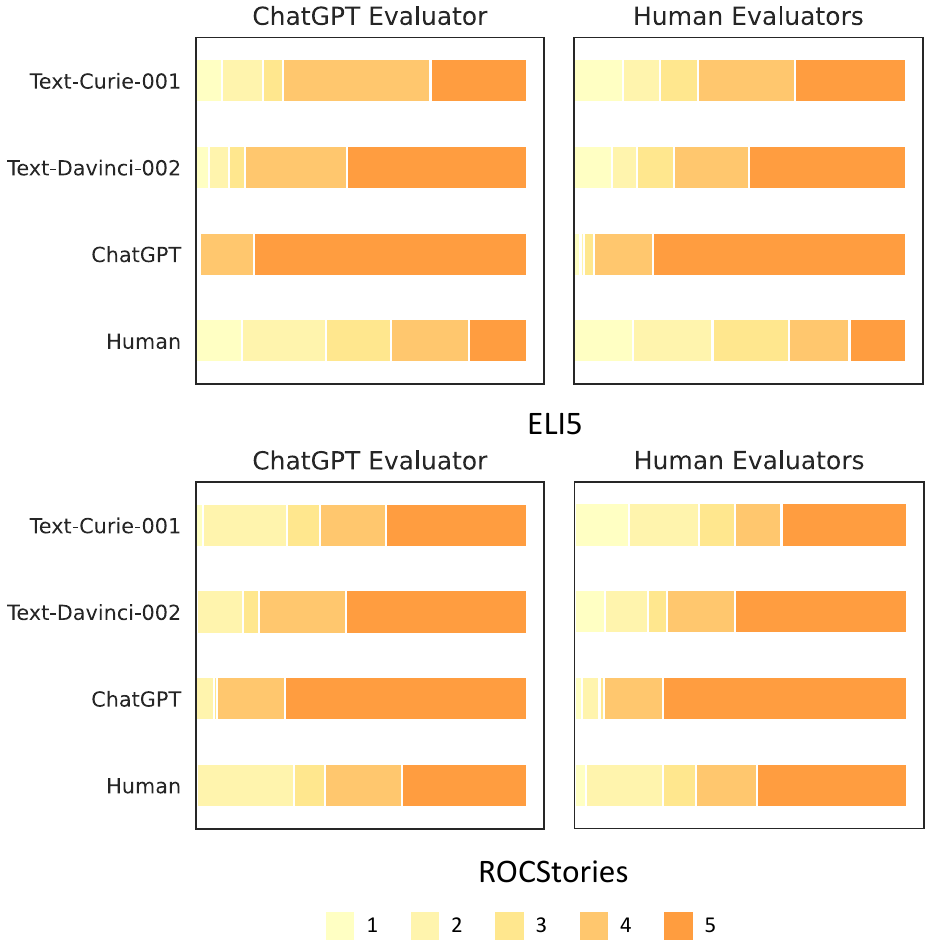}
    \caption{The distribution of scores assigned by LLM and human evaluators for generations (ELI5 and ROCStories) written by humans and models with different qualities. The score (1 to 5) ratios across different sources are distinct, suggesting that both LLMs and humans can discern these differences.}
    \label{fig:quality_app}
\end{figure}

\paragraph{Evaluation of Samples with Varied Quality Levels}
We evaluate sentences from both humans and models with varying quality, expecting that differences between generative sources will be reflected in the evaluation results. 
The comparison of the scoring patterns of the LLM and human evaluators on the Self-Instruct task is presented in Figure~\ref{fig:score_of_diff_qualities}. 
Figure~\ref{fig:quality_app} presents the evaluation score distributions of LLM and humans to distinguish generations from different sources (models and humans) on ELI5 and ROCStories tasks.
The LLM tends to assign more positive scores than humans, reflected by the smaller range of low scores.

\section{Details of Human Evaluations}
\label{apx:expert_select}

\paragraph{Human Expert Selection}
In this paper, a total of 7 NLP researchers and 15 crowdsource annotators participated in the criterion scrutiny.
The laypeople were hired through a qualifying exam. 
NLP researchers, due to their familiarity with the evaluated tasks, can be considered experts in evaluating criteria for providing valuable insights into the suitability of the criteria.
Including both researchers and laypeople in this stage ensures that the evaluation considers both the scientific theories of the NLP tasks and the preferences of normal users.

\paragraph{Annotator Compensation}
On average, evaluators spent approximately five minutes on task criteria establishment. 
We compensate evaluators \$2.5 per task. They take around six minutes to complete a single instance evaluation, which involves assessing five to six criteria. 
We pay them US\$35.5 an hour. The local minimum salary in the year 2023 is US\$15.5 per hour for part-time jobs. The annotation does not involve any personally sensitive information.
Evaluators tend to slow down their evaluation speed in the middle of the evaluation process, which can affect time calculations. To ensure the accuracy of our time calculations and the quality of the annotations, we periodically check the annotator's results every few batches. This helps ensure that the quality of the annotations and the median time taken per annotator are consistent with our pay rate.

\paragraph{Quality Control}
In Section~\ref{sec:criteria_gen}, NLP researchers who possess familiarity with the evaluated tasks are recognized as the evaluation experts responsible for assessing the quality of the evaluation criteria proposed by the LLM.
During the evaluation process based on predefined criteria (Section~\ref{sec:instance_eval}), we engage crowd annotators to participate in scoring and refining the evaluations provided by the LLM. 
Annotators must complete a qualifying exam before evaluating:
(1) They are first pre-screened with a qualification study, which involves reading an evaluation guideline and evaluating three instances from three datasets. 
(2) We individually review the submitted evaluations from the qualification study and provide feedback to clarify any misconceptions about the task.
(3) Evaluators who performed well on the qualification study and demonstrated a thorough understanding of the evaluation guidelines are selected to participate in the human evaluation. 
(4) Throughout the whole process, we maintain constant communication with evaluators to answer any questions.
Ultimately, we selected 15 native speakers (5 evaluators per task) from North America as human annotators.

\begin{figure*}
    \centering
    \includegraphics[width=\linewidth]{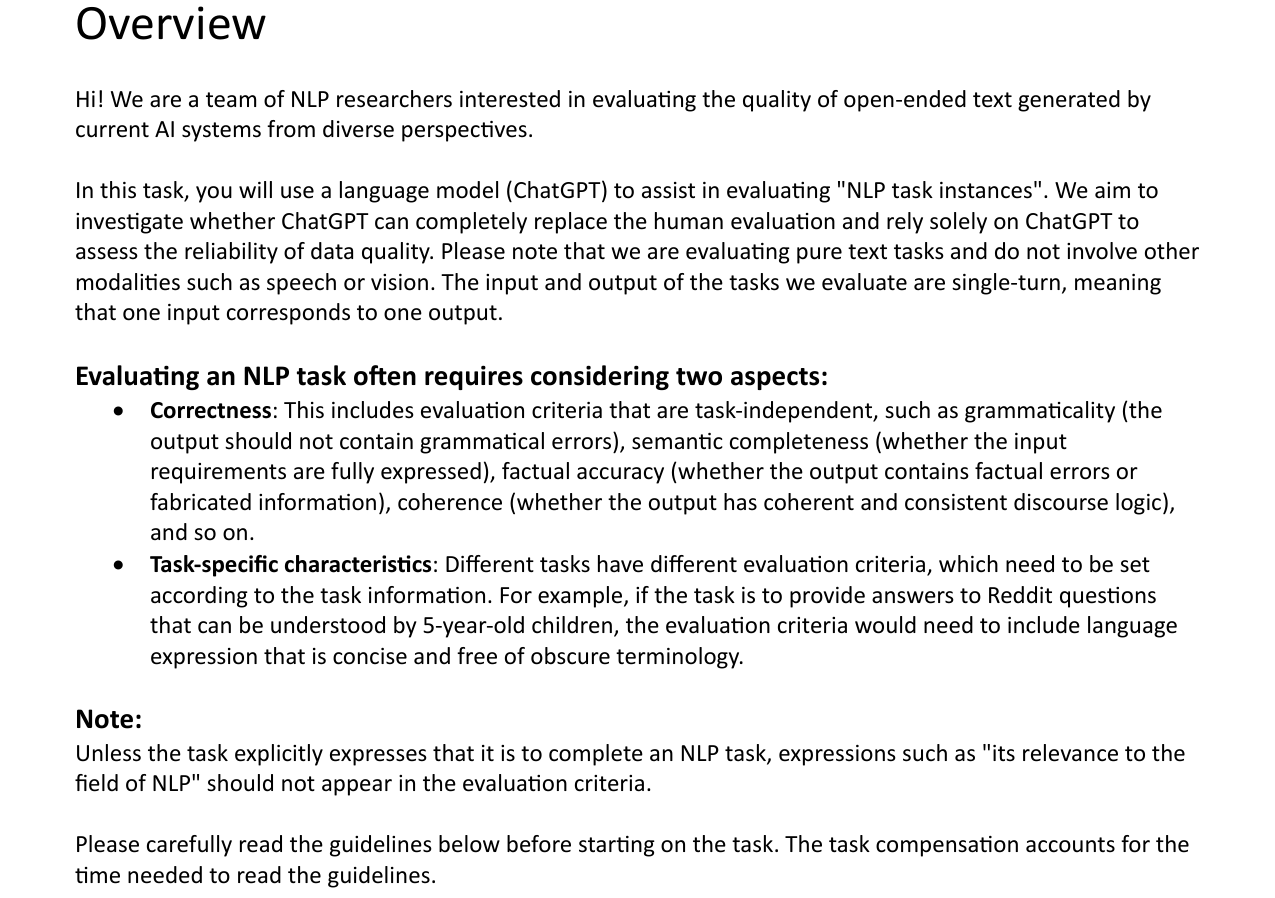}
    \caption{The \textbf{first} page of the evaluation guideline, which is used in the qualification test.}
    \label{fig:guide_1}
\end{figure*}

\begin{figure*}
    \centering
    \includegraphics[width=\linewidth]{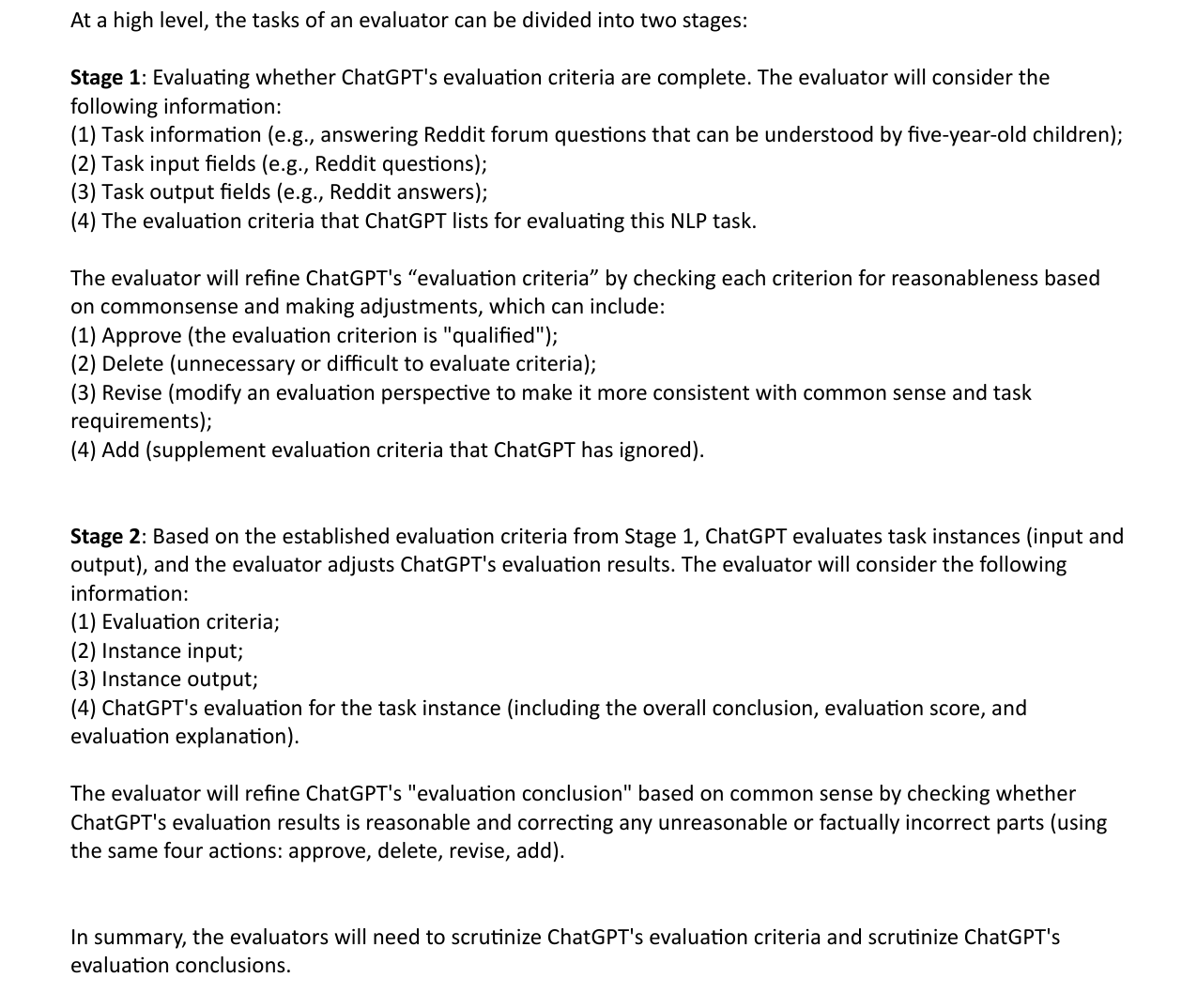}
    \caption{The \textbf{second} page of the evaluation guideline, which is used in the qualification test.}
    \label{fig:guide_2}
\end{figure*}

\paragraph{Annotation Guidelines}
\label{apx:anno_guide}
Figure~\ref{fig:guide_1} and Figure \ref{fig:guide_2} show the evaluation guidelines we used for the whole evaluation pipeline. 
We ask crowd evaluators to read these guidelines as part of the qualification study. 
Only evaluators who demonstrated a thorough understanding of the guidelines and tasks were permitted to participate in the main round of the evaluation procedure.

\section{Evaluation Platform}
\label{apx:platform}
\begin{figure*}
    \centering
    \includegraphics[width=\linewidth]{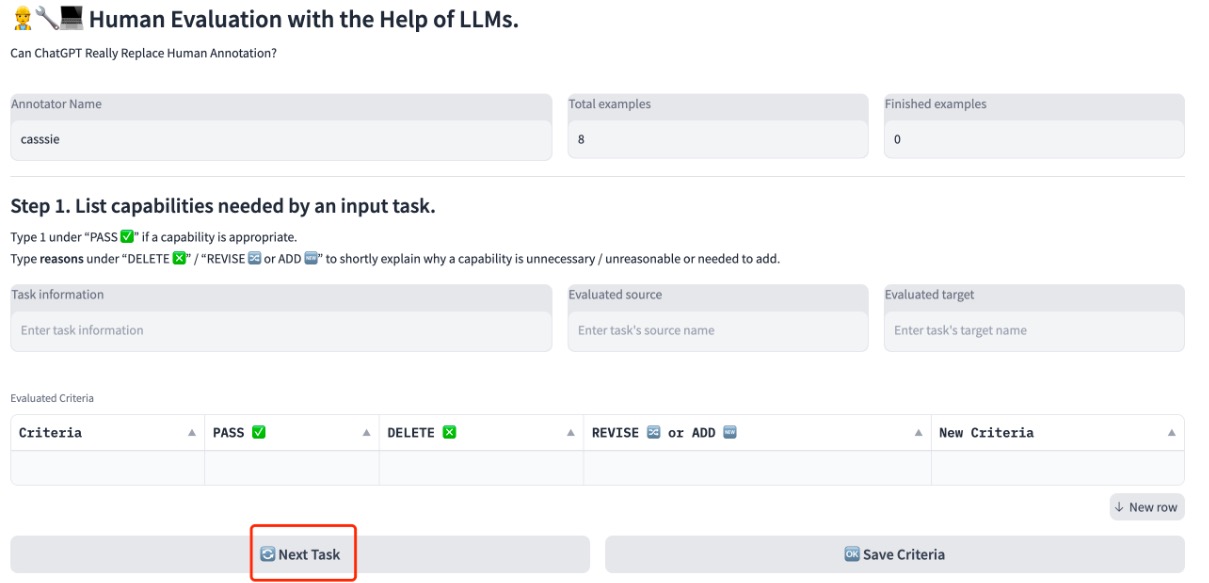}
    \caption{The upper part of the evaluation interface in criteria establishment.}
    \label{fig:demo_1}
\end{figure*}

\begin{figure*}
    \centering
    \includegraphics[width=\linewidth]{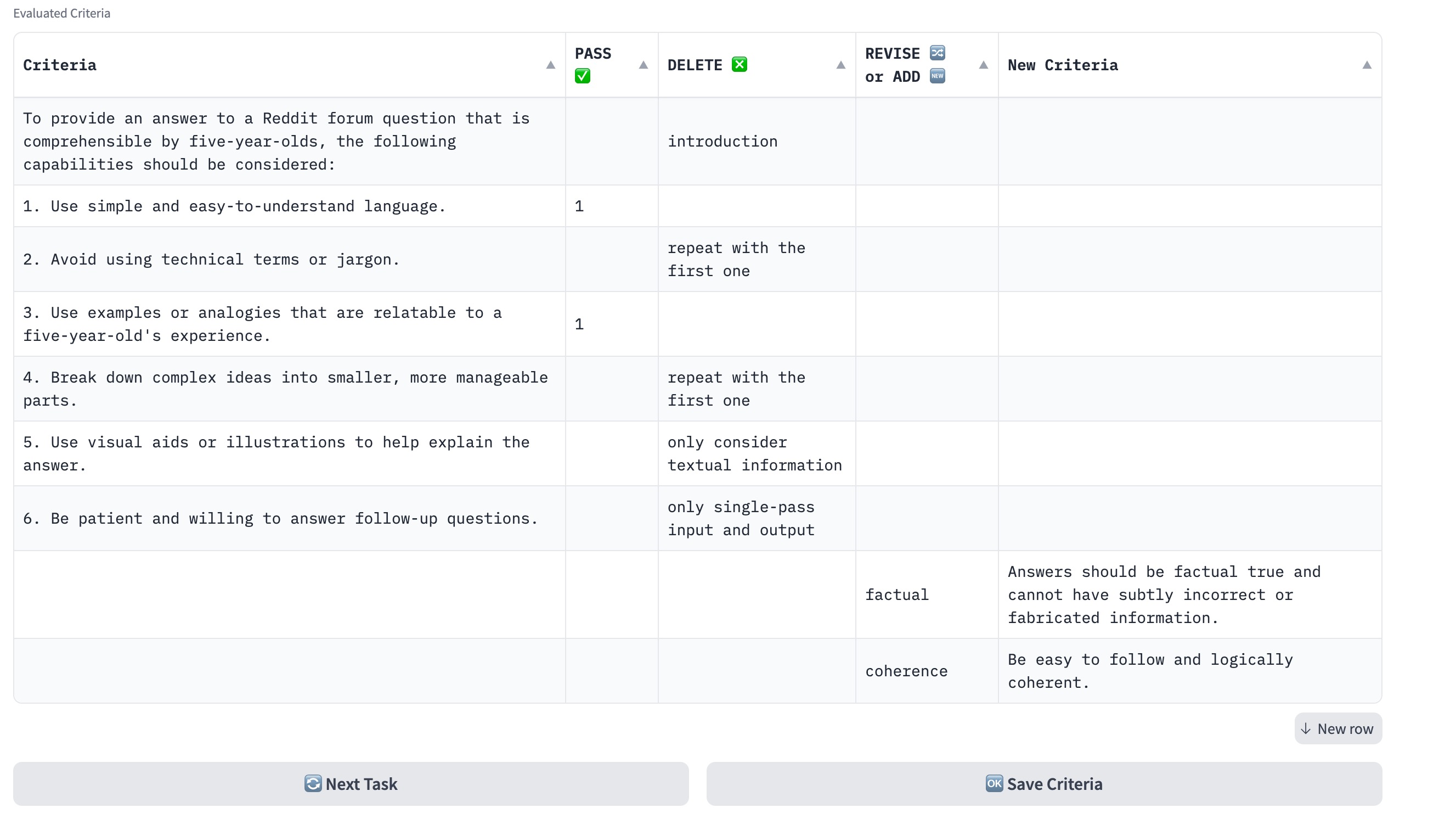}
    \caption{The lower part of the evaluation interface in criteria establishment.}
    \label{fig:demo_2}
\end{figure*}

\begin{figure*}
    \centering
    \includegraphics[width=\linewidth]{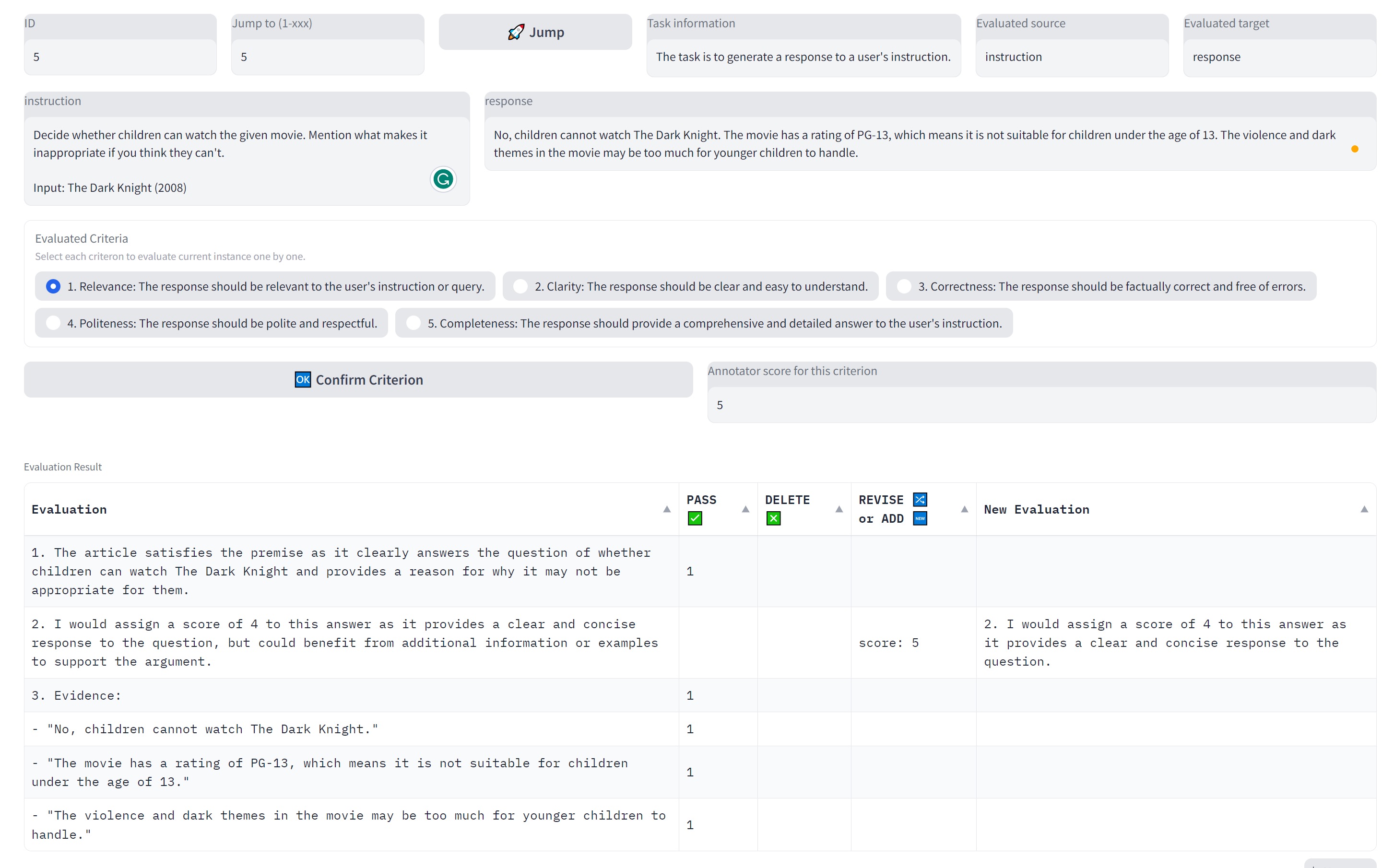}
    \caption{The evaluation interface in instance-level evaluation.}
    \label{fig:demo_3}
\end{figure*}

We build our platform using Gradio repository\footnote{\url{https://gradio.app/}} and display the screenshots of the evaluation pipeline in Figure~\ref{fig:demo_1}-\ref{fig:demo_3}.

\begin{table*}[htb!]
\centering
\small
\begin{tabular}{lp{0.30\textwidth}p{0.55\textwidth}}
\toprule
\textbf{Task} & \textbf{Task Description} & \textbf{Finalized Evaluation Criteria} \\
\midrule
ELI5 & ELI5 is a task for long-form question answering. It contains complex, diverse questions that require explanatory multi-sentence answers. This task aims to provide an explanatory answer that is comprehensible to five-year-olds. & 1. Comprehensibility: The answers should be written in simple and clear language that a five-year-old can understand. \newline 2. Accuracy: The answers should be factually accurate and provide correct explanations. \newline 3. Coherence: The answers should be well-structured and coherent, with a logical flow of information. \newline 4. Engagement: The answers should be engaging and interesting for a five-year-old. \newline 5. Use of Examples and Analogies: The answers should incorporate relevant examples or analogies to aid comprehension.  \\
\midrule 
ROCStories & ROCStories is a task for commonsense short story generation. The task aims to generate stories that contain a variety of commonsense causal and temporal relations between everyday events. & 1. Relevance: The story should demonstrate an understanding of the context and background information provided in the prompt. It should incorporate relevant details about the given prompt or topic.  \newline 2. Coherence: The generated story should be coherent and make logical sense. The events and actions should be connected in a meaningful way, following a clear causal and temporal progression. \newline 3. Clarity: The generated story should be easily understandable, with proper grammar, syntax, and vocabulary. \newline  4. Commonsense Knowledge: The story should effectively utilize and demonstrate correct commonsense knowledge in its narrative. \newline 5. Creativity: The story should provide a fresh and interesting perspective on the given topic or prompt. \newline 6. Length: The story should adhere to the specified length and structure requirements. \\
\midrule 
GSM8K & GSM8K is a task for grade school math word problem-solving. These problems take between 2 and 8 steps to solve, and solutions primarily involve performing a sequence of elementary calculations using basic arithmetic operations ($+$ $-$ $\times$ $\div$) to reach the final answer. The task aims to generate a solution chain that demonstrates logical and valid reasoning and calculation.  & 1. Logical Reasoning: The solution chain should demonstrate a logical and valid sequence of steps to reach the final answer. \newline 2. Numerical Understanding: The model should accurately perform the necessary calculations and use the correct mathematical operations ($+$ $-$ $\times$ $\div$) to solve the problem. \newline 3. Completeness: The solution chain should include all the necessary steps and calculations required to solve the problem.\\  
\bottomrule
\end{tabular}
    \caption{The evaluated tasks, i.e., ELI5, ROCStories, and GSM8K, along with their respective criteria for sample-wise evaluation. The criteria initially proposed by the LLM are subsequently examined and validated by human experts.}
    \label{app:criteria_details} 
\end{table*}

\begin{table*}[htb!]
\centering
\small
\begin{tabular}{lp{0.30\textwidth}p{0.45\textwidth}}
\toprule
\textbf{Task} & \textbf{Task Description} & \textbf{Finalized Evaluation Criteria} \\
\midrule
Self-Instruct (\fontsize{7pt}{12pt}\selectfont\textit{Twitter}) & The task is to generate content intended for social media platforms. 
 & 1. Relevance: The generated content should address the given task and provide appropriate information. \newline 2. Coherence: The generated content should flow naturally. \newline 3. Tone and Style: The generated content should match the specified tone and style requirements, such as casual, professional, or formal. \newline 4. Engagement: The generated content should be engaging and capture the attention of the target audience, such as asking for responses or feedback. \newline 5. Cultural Sensitivity: The generated content should be culturally sensitive and avoid any offensive or inappropriate language.\\
\midrule
Self-Instruct (\fontsize{7pt}{12pt}\selectfont\textit{IMDB}) & The task is to generate responses for instructions within the movie domain. & 1. Relevance: The responses should directly answer the given instructions and provide accurate information. \newline 2. Coherence: The responses should have a clear structure and organization, presenting the information in a logical and easy-to-follow manner. \newline 3. Accuracy of movie information: The responses should accurately describe the movie, including its rating and content. \newline 4. Creativity: The responses should show creativity in how they describe the movie, using interesting and attention-grabbing language.\\
\midrule
Self-Instruct (\fontsize{7pt}{12pt}\selectfont\textit{Gmail}) & The task is to write an email based on an instruction. & 1. Language: The email should be written with proper grammar, spelling, and punctuation. \newline 2. Coherence: The email should be well-organized and easy to understand. The main points should be clearly stated and supported with relevant information. \newline 3. Relevance: The email should address the specific situation and follow the given instructions or requirements. \newline 4. Appropriate greetings and closings: The email should use appropriate greetings and closings, such as "Dear [name]" and "Sincerely," to maintain a professional tone. \newline 5. Appropriate tone and style: The email should use an appropriate tone and style for the situation, whether it is formal, informal, friendly, or professional.\\
\midrule
Self-Instruct (\fontsize{7pt}{12pt}\selectfont\textit{Notion}) & The task is to create a plan or outline based on a given instruction.  & 1. Accuracy: The plan should accurately reflect the given information and instructions. \newline 2. Organization: The plan should be well-organized, with a logical flow and structure in a clear and readable format. \newline 3. Completeness: The plan should cover all the necessary tasks or elements mentioned in the given information.\\
\midrule
Self-Instruct (\fontsize{7pt}{12pt}\selectfont\textit{Tasty}) & The task is to generate responses for instructions related to food.  & 1. Relevance: The responses should directly address the given instruction and provide relevant information. \newline 2. Organization: The responses should be well-structured and organized, presenting the information in a logical and easy-to-follow manner. \newline 3. Domain Knowledge: The responses should demonstrate a good understanding of food-related concepts, ingredients, cooking techniques, and culinary practices.\\
\bottomrule
\end{tabular}
    \caption{The evaluated dataset, Self-Instruct, comprises a range of daily instructions for various scenarios, each requiring slightly different criteria for evaluation.}
    \label{exp:eval_detail_self_instruct} 
\end{table*}

\end{document}